\documentclass{article}
\usepackage{ijcai23}

\usepackage{times}
\usepackage{soul}
\usepackage{url}
\usepackage[hidelinks]{hyperref}
\usepackage[utf8]{inputenc}
\usepackage[small]{caption}
\usepackage{graphicx}
\usepackage{amsmath}
\usepackage{amsthm}
\usepackage{booktabs}
\usepackage{algorithm}
\usepackage{algorithmic}
\usepackage[switch]{lineno}

\usepackage{amsfonts}
\usepackage{subfigure}
\usepackage{multirow}
\usepackage[table]{xcolor}
\usepackage{arydshln}
\usepackage{tabularx}
\usepackage{dsfont}

\usepackage{txfonts}

\definecolor{cred}{HTML}{A10035}
\definecolor{cyellow}{HTML}{FEC260}
\definecolor{cgreen}{HTML}{3FA796}
\definecolor{cpurple}{HTML}{2A0944}
\definecolor{ggray}{RGB}{127,127,127}
\definecolor{aliceblue}{rgb}{0.94, 0.97, 1.0}
\definecolor{convnext_purple}{HTML}{483A7F}
\definecolor{convnext_yellow}{HTML}{E99675}

\newcommand{\pub}[1]{\color{gray}{\tiny{#1}}}
\newcommand{\bsl}[1]{\color{ggray}{\scriptsize{{#1}}}}

\newcommand{\thickhline}{%
	\noalign {\ifnum 0=`}\fi \hrule height 1pt
	\futurelet \reserved@a \@xhline
}

\newcommand{\increase}[1]{
	\fontsize{6pt}{0.5em}\selectfont\color{purple}{$\uparrow$~\textbf{#1}}
}
\newcommand{\decrease}[1]{
	\fontsize{6pt}{0.5em}\selectfont\color{gray!48}{$\downarrow$~\textbf{#1}}
}

\newcolumntype{x}[1]{>{\centering\arraybackslash}p{#1pt}}
\newcolumntype{y}[1]{>{\raggedright\arraybackslash}p{#1pt}}
\newcolumntype{z}[1]{>{\raggedleft\arraybackslash}p{#1pt}}

\title{WiCo: Win-win Cooperation of Bottom-up and Top-down \\Referring Image Segmentation}

\author{
Zesen Cheng$^{1,2}$\thanks{$^\dagger$\ Corresponding Author}, Peng Jin$^{1,2}$, Hao Li$^{1,2}$, Kehan Li$^{1,2}$, Siheng Li$^{4}$\\
Xiangyang Ji$^{4}$, Chang Liu$^{4}$~$^\dagger$ {\normalfont and} Jie Chen$^{1,2,3}$~$^\dagger$ \\
\affiliations
{$^{1}$ School of Electronic and Computer Engineering, Peking University, Shenzhen, China} \\
{$^{2}$ AI for Science (AI4S)-Preferred Program, Peking University Shenzhen Graduate School, China} \\
{$^{3}$ Peng Cheng Laboratory, Shenzhen, China} \\
{$^{4}$ Tsinghua University, Beijing, China} \\ 
\emails
{\{cyanlaser, jp21, kehanli\}@stu.pku.edu.cn},
{lisiheng21@mails.tsinghua.edu.cn}\\
{\{xyji, liuchang2022\}@tsinghua.edu.cn},
{\{lihao1984, jiechen2019\}@pku.edu.cn}
}
\renewcommand\footnotemark{}

\begin{document}

\maketitle
\begin{abstract}
The top-down and bottom-up methods are two mainstreams of referring segmentation, while both 
 methods have their own intrinsic weaknesses.
Top-down methods are chiefly disturbed by \textbf{Polar Negative}~(PN) errors owing to the lack of fine-grained cross-modal alignment.
Bottom-up methods are mainly perturbed by \textbf{Inferior Positive}~(IP) errors due to the lack of prior object information.
Nevertheless, we discover that two types of methods are highly complementary for restraining respective weaknesses but the direct average combination leads to harmful interference.
In this context, we build \underline{\textbf{Wi}}n-win \underline{\textbf{Co}}operation~(WiCo) to exploit complementary nature of two types of methods on both interaction and integration aspects for achieving a win-win improvement.
For the interaction aspect, \textbf{Complementary Feature Interaction}~(CFI) provides fine-grained information to top-down branch and introduces prior object information to bottom-up branch for complementary feature enhancement.
For the integration aspect, \textbf{Gaussian Scoring Integration}~(GSI) models the gaussian performance distributions of two branches and weighted integrates results by sampling confident scores from the distributions.
With our WiCo, several prominent top-down and bottom-up combinations achieve remarkable improvements on three common datasets with reasonable extra costs, which justifies effectiveness and generality of our method.
\end{abstract}

\begin{figure}[t]
\centering
\includegraphics[width=0.48\textwidth]{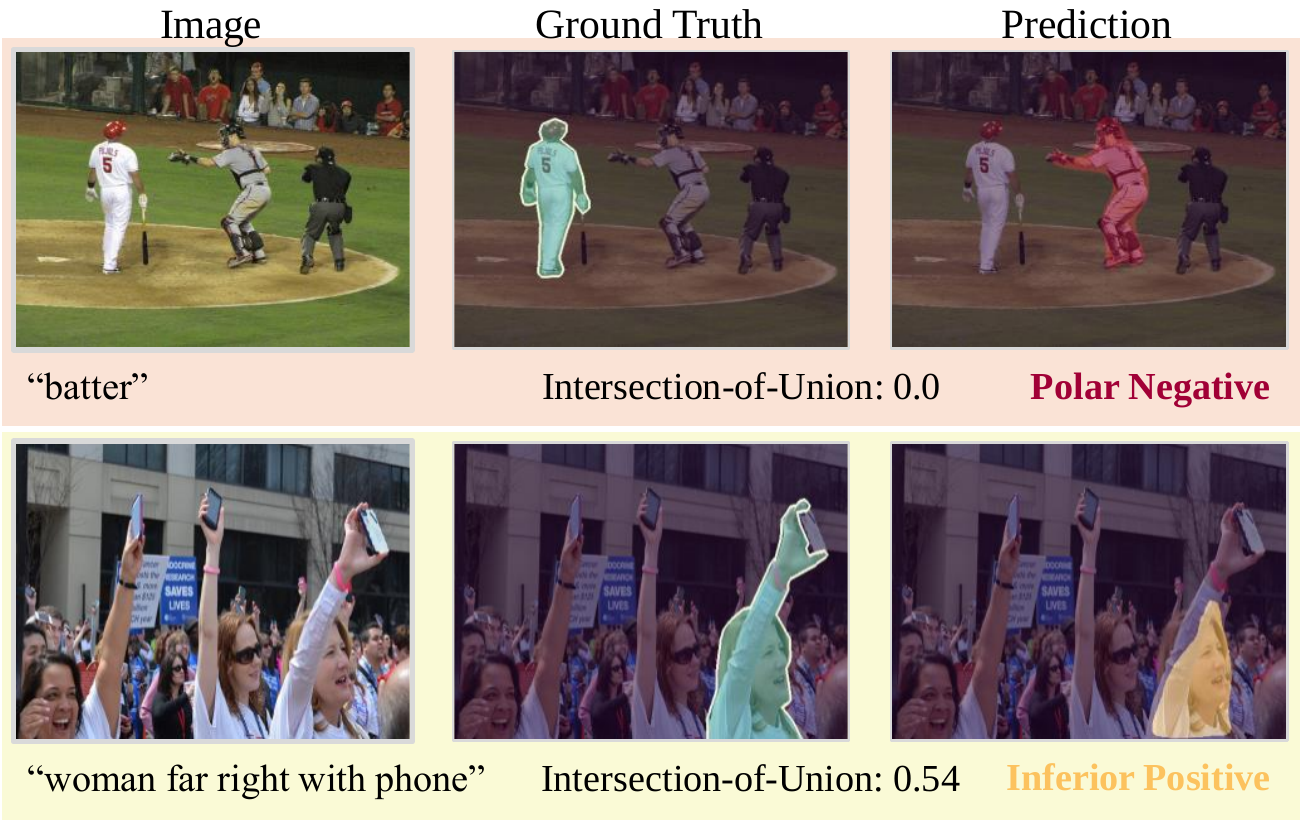}
\caption{\textbf{Visualization of some failure cases}. The sample of first row is defined as \textbf{\color{cred}{Polar Negative}} samples which have no overlap with the \textbf{\color{cgreen}{Ground Truth}} and the Intersection-of-Union~(IoU) closes to 0. The sample of second row is defined as \textbf{\color{cyellow}{Inferior Positive}} samples which ignore some object parts and IoU ranges from 0.5 to 0.8. Existing methods still fail to process these two types of errors.}
\label{fig:motivation}
\end{figure}
\begin{figure*}[t]
\centering
\subfigtopskip=0pt    
\subfigbottomskip=1pt 
\subfigcapskip=-2pt   
\begin{minipage}[b]{0.24\textwidth}
    \subfigure[Top-down]{
        \centering
        \includegraphics[width=0.97\textwidth]{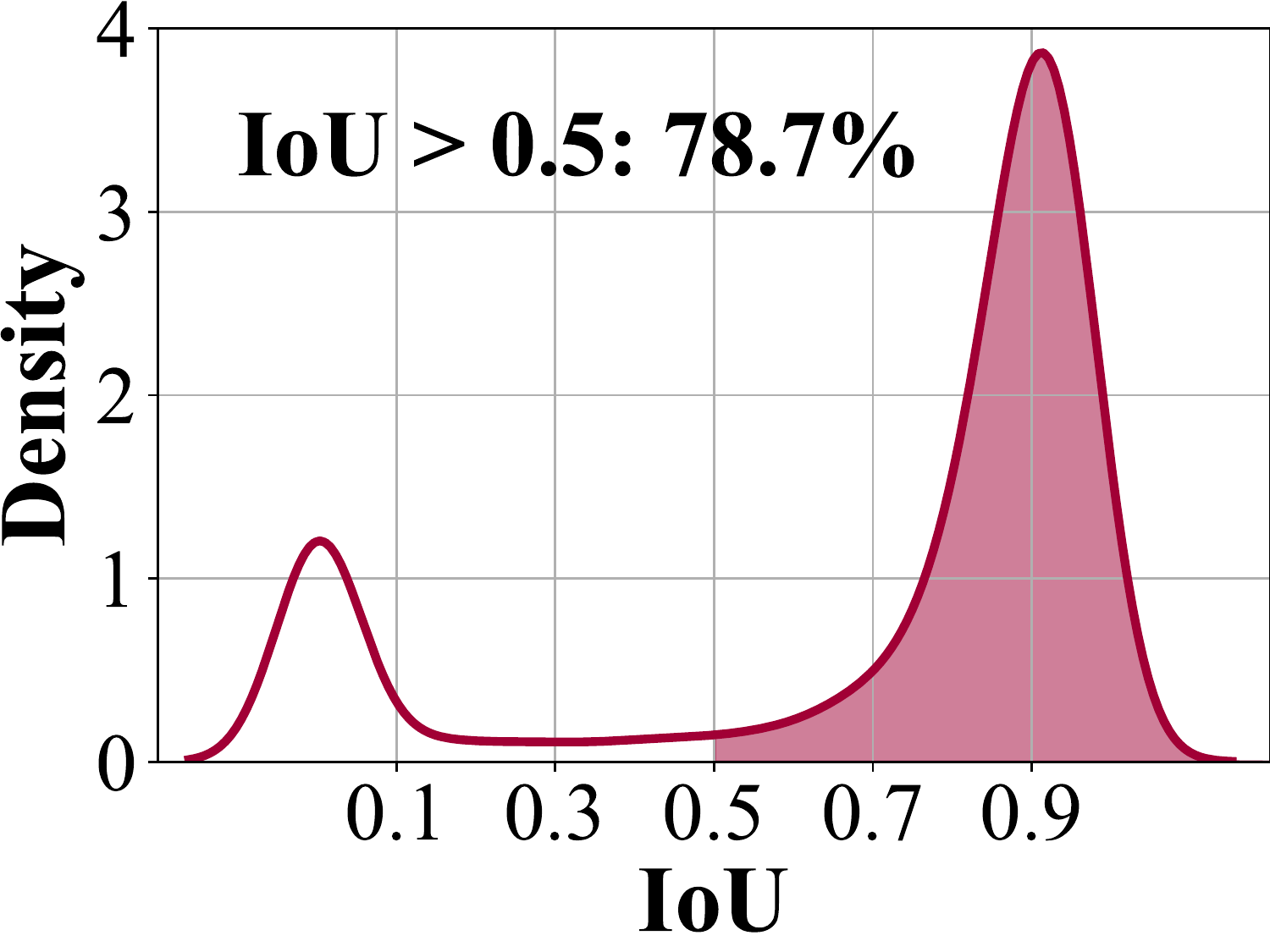}
    }
\end{minipage}
\hfill
\begin{minipage}[b]{0.24\textwidth}
    \subfigure[Bottom-up]{
        \centering
        \includegraphics[width=0.97\textwidth]{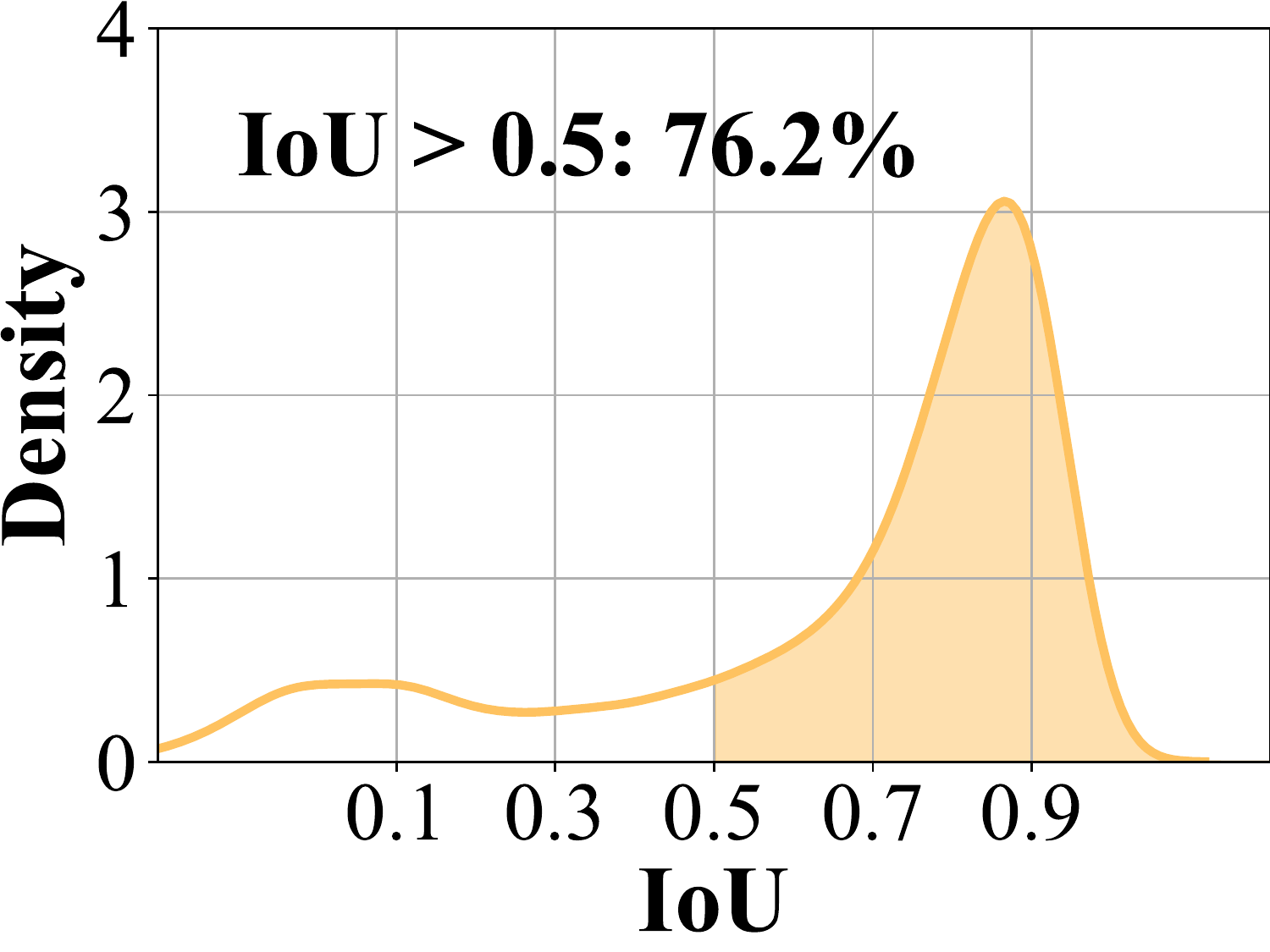}
    }
\end{minipage}
\hfill
\begin{minipage}[b]{0.24\textwidth}
    \subfigure[Direct combination]{
        \centering
        \includegraphics[width=0.97\textwidth]{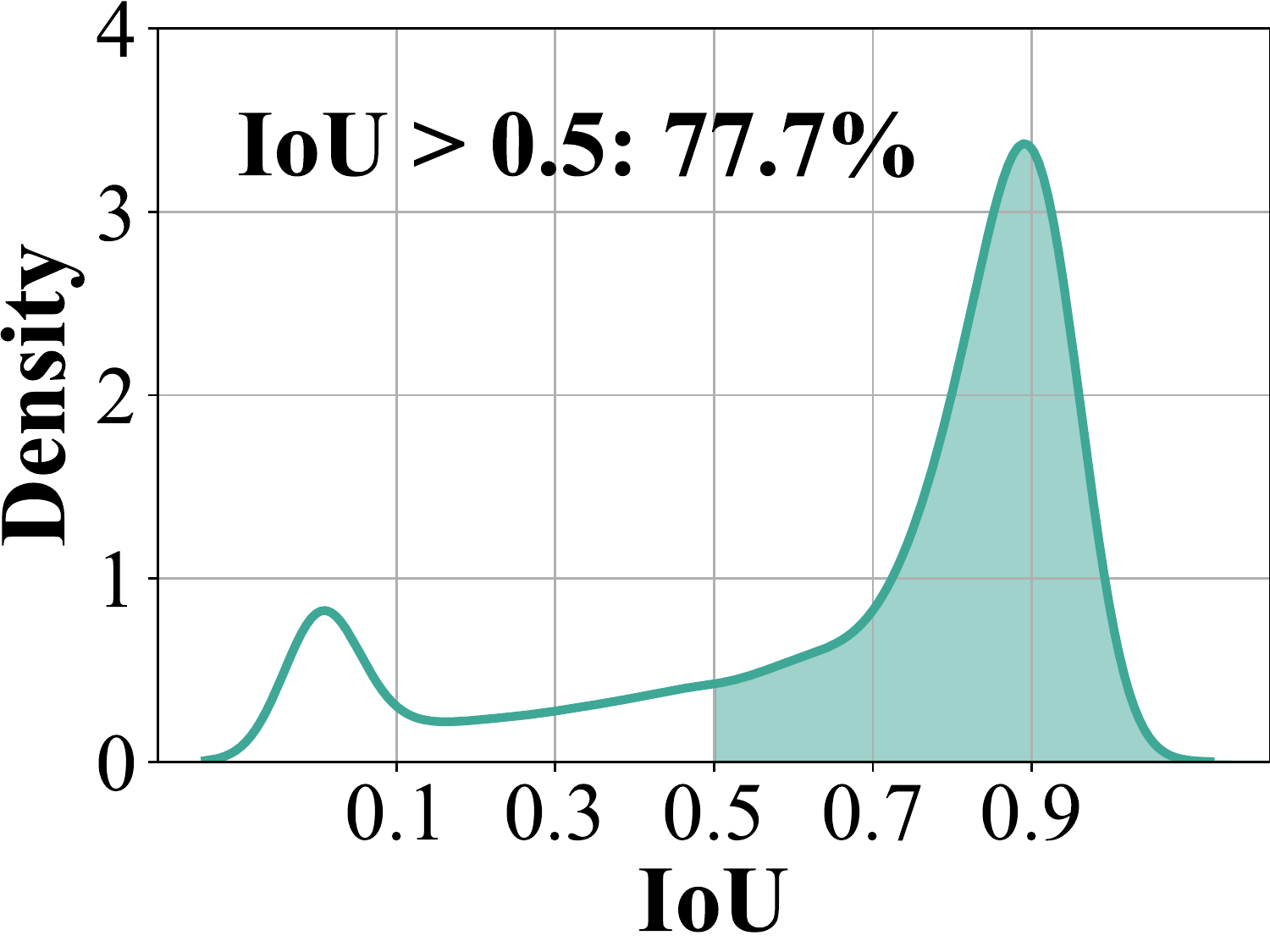}
    }
\end{minipage}
\hfill
\begin{minipage}[b]{0.24\textwidth}
    \subfigure[WiCo]{
        \centering
        \includegraphics[width=0.97\textwidth]{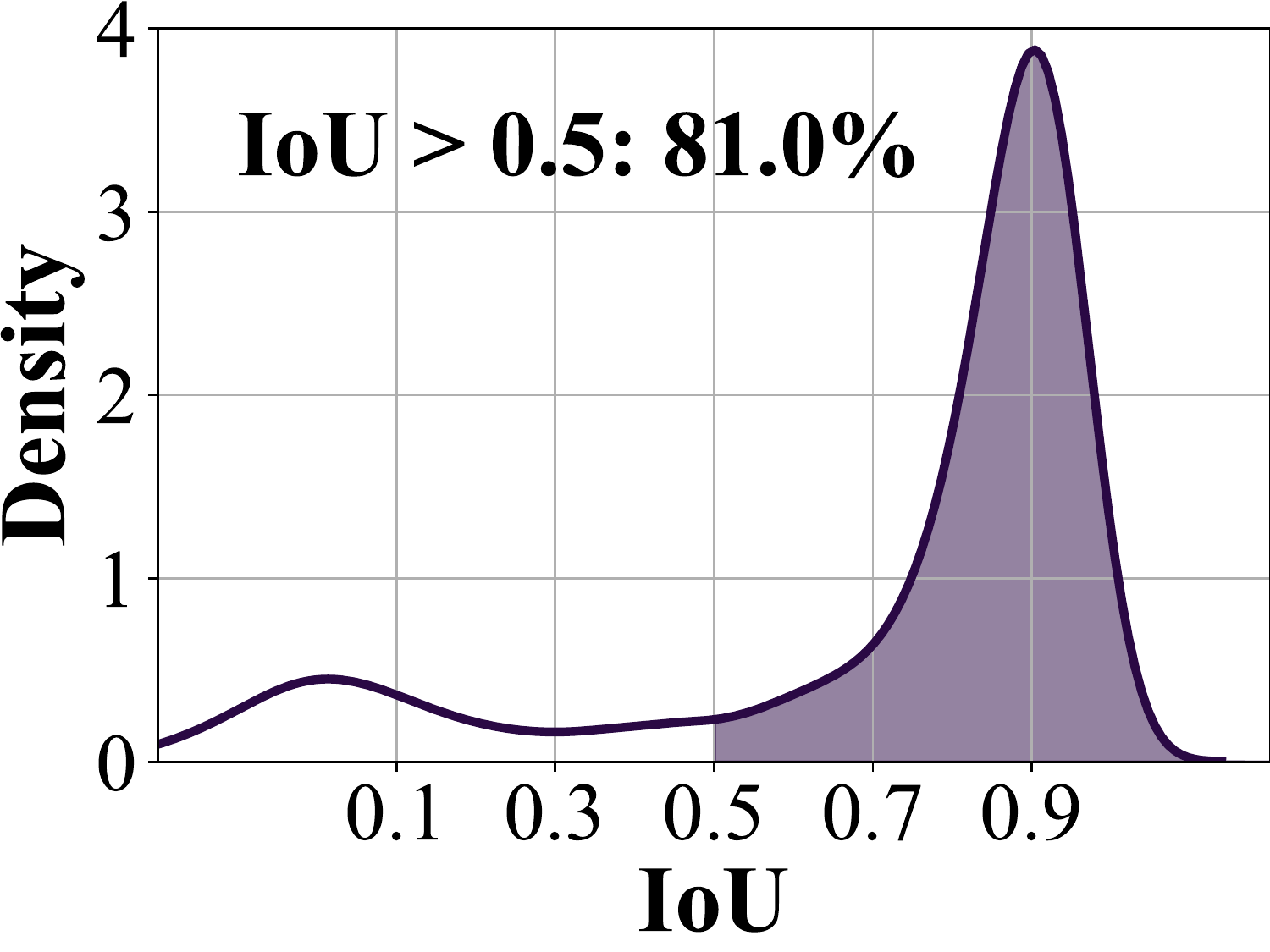}
    }
\end{minipage} 
\caption{\textbf{The IoU distribution} of (a) top-down method (\textbf{\color{cred}{$ 62.62 $ IoU}}), (b) bottom-up method (\textbf{\color{cyellow}{$ 65.65 $ IoU}}), (c) direct combination between top-down and bottom-up methods (\textbf{\color{cgreen}{$ 68.55 $ IoU}}) and (d) our WiCo (\textbf{\color{cpurple!60}{$ 71.74 $ IoU}}). We discover that the valid scope of the two methods is highly complementary. However, the direct combination leads to adverse cooperation. Our proposed WiCo mechanism adequately absorbs the advantages for better coping with those failure cases. These distributions are calculated on the RefCOCO val split.}
\label{fig:intuition}
\end{figure*}
\section{Introduction}
\label{sec:intro}

Referring image segmentation~(RIS) is a new type of segmentation task aiming to segment the object referred by a natural query expression. 
The current approaches for referring image segmentation can be broadly classified into two categories~\cite{hui2020linguistic}, i.e., top-down and bottom-up methods. 
Top-down methods calculate the object-centric cross-modal alignment between each region proposal from pretrained detector and query for getting cross-modal instance embeddings and then decode cross-modal instance embeddings to alignment score for retrieving the most confident region proposal as segmentation result~\cite{yu2018mattnet,liu2019learning}.
Bottom-up methods calculate the fine-grained cross-modal alignment between each pixel and query for acquiring cross-modal pixel embeddings and then decode the embeddings to retrieve those pixels of referred object~\cite{wang2022cris,yang2022lavt}.
However, according to our observations in Figure~{\color{red}{\ref{fig:motivation}}}, existing top-down and bottom-up methods are still perturbed by two types of errors: \textbf{Polar Negative}~(PN) and \textbf{Inferior Positive}~(IP). These two errors can be identified by the Intersection-over-Union~(IoU) between predictions and ground truths. PN samples are those predictions that have nearly no overlap with the ground truth (IoU $\rightarrow$ 0). IP samples are those predictions that ignore some components of the referred object (IoU $\in [0.5, 0.8]$).

To analyze how top-down and bottom-up methods are disturbed by PN errors and IP errors, we visualize the IoU distribution of top-down and bottom-up methods in Figure~{\color{red}{\ref{fig:intuition}}}. We split the distribution curve into two parts: positive set~(samples with IoU $ > $ 0.5) and negative set~(samples with IoU $ < $ 0.5). 
As shown in Figure~{\color{red}{\ref{fig:intuition}}} (a), the positive set of top-down method achieves higher location precision than bottom-up method because the prior object information suppresses low-quality IP samples. But top-down method easily generates PN samples on negative samples due to the lack of fine-grained cross-modal alignment.
In Figure~{\color{red}{\ref{fig:intuition}}} (b), bottom-up method outputs a large number of IP samples on positive set owing to the lack of prior object information. Nevertheless, the negative set of bottom-up methods has smoother distribution than top-down methods and the PN samples of bottom-up methods are far fewer than top-down methods because fine-grained information provides robust cross-modal alignment.
According to the analysis above, we can conclude that top-down and bottom-up methods are highly complementary.

Intuitively, we can fuse top-down and bottom-up methods by direct combination, i.e., straightly averaging their results. However, as shown in Figure~{\color{red}{\ref{fig:intuition}}} (c), this scheme leads to harmful cooperation between top-down and bottom-up methods which can be attributed to the lack of feature interaction and the inappropriate integration of results.
In this context, we build \underline{\textbf{Wi}}n-win \underline{\textbf{Co}}opEration~(WiCo) to exploit the complementary nature of top-down and bottom-up branches by interacting with each other and integrating results of two branches in an adaptive manner, which follows ``Interaction then Integration" paradigm for compensating the defect of direct combination.
WiCo contains two modules: \textbf{Complementary Feature Interaction}~(CFI) and \textbf{Gaussian Scoring Integration}~(GSI).
CFI is designed to perform interaction between features of two branches for compensating the lack of fine-grained information in top-down branch and prior object information in bottom-up branch.
GSI is designed to model the gaussian performance distributions of top-down and bottom-up branches and adaptively integrate results of two branches by sampling confidence scores from the distributions. 
Figure~{\color{red}{\ref{fig:intuition}}} (d) shows that our framework largely reduces IP errors and PN errors and generates fine IoU distribution with the merits of top-down and bottom-up methods, which demonstrates our method is more effective than direct combination scheme for incorporating top-down and bottom-up methods.

In summary, the main contributions are as follows:
\begin{itemize} 
\item We analyze the behavior of several top-down methods and bottom-up methods when facing PN and IP errors. According to the analysis, we discover that existing top-down and bottom-up methods are highly complementary in how to cope with PN errors and IP errors.
\item We propose WiCo to adequately exploit the characteristics of top-down and bottom-up methods methods and let them effectively complement each other on both interaction and integration aspects, which can better process PN errors and IP errors than intuitive direct combination. 
\item 
Extensive experiments show that our WiCo can boost the performance of top-down and bottom-up methods methods by 2.25\%$\sim$6.66\% under three common datasets: \textit{RefCOCO}, \textit{RefCOCO+} and \textit{G-Ref} with reasonable cost.
\end{itemize}

\begin{figure*}[ht]
\centering
\includegraphics[width=1.0\textwidth]{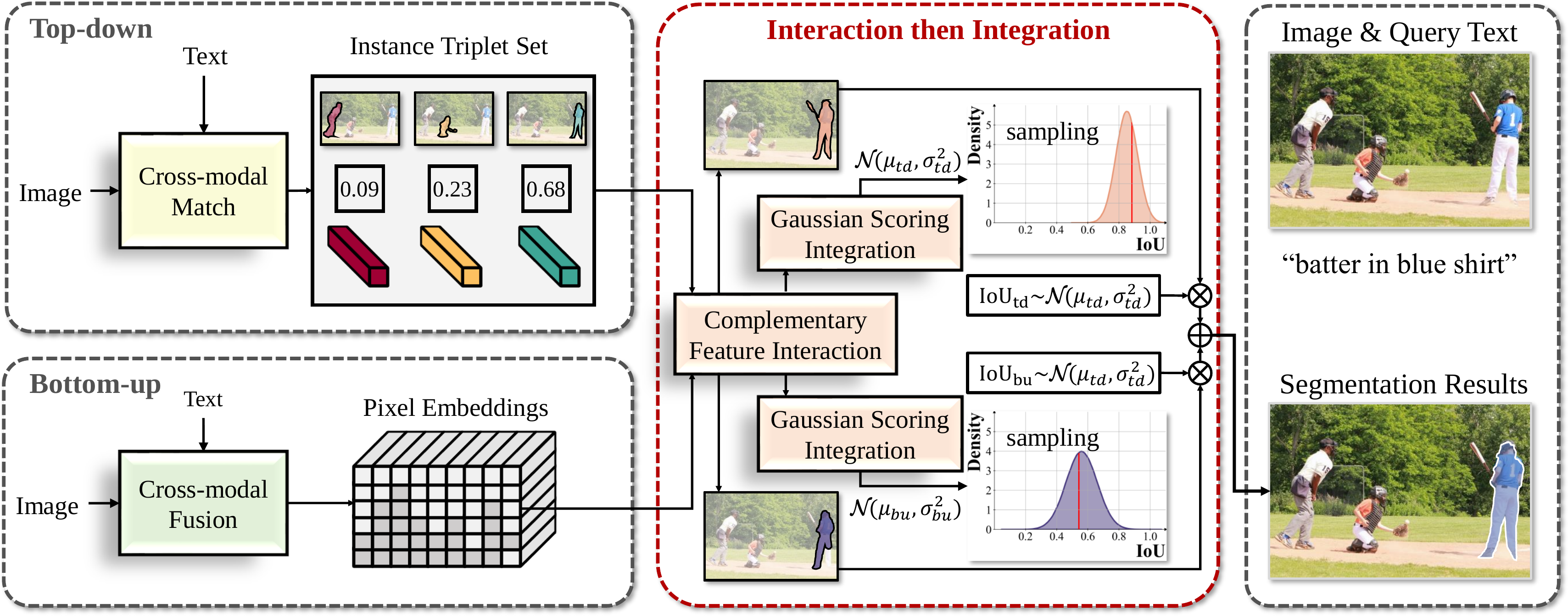} 
\caption{\textbf{The overall pipeline of our WiCo.} Firstly, top-down and bottom-up branches acquire the respective features and results. Then, these features and results are input into CFI~(Complementary Feature Interaction) for knowledge interaction. Finally, we use GSI~(Gaussian Scoring Integration) to predict the performance distributions of two branches and weighted integrate the results of two branches according to the confidence score sampled from the performance distributions. The modules inside the \textbf{\color{cred}{red dashed box}} are our main contribution.}
\label{fig:framework}
\end{figure*}
\section{Related Work}

\paragraph{Top-down Method.} Previous efforts on top-down style referring image segmentation are about how to calculate better object-centric cross-modal alignment between region proposals of instances and referring expression query. For example, MAttNet~\cite{yu2018mattnet} decomposes referring expressions into three components to match instances. NMTree~\cite{liu2019learning} regularizes the cross-modal alignment along the dependency parsing tree of the sentence. CAC~\cite{chen2019referring} introduces cycle consistency between referring expression and its reconstructed caption into the reasoning part of network for boosting cross-modal alignment.

\paragraph{Bottom-up Method.} Previous efforts on bottom-up style referring image segmentation mainly focus on densely aligning and fusing visual and linguistic features for better cross-modal pixel features. For example, early works~\cite{hu2016segmentation,li2018referring} propose to use simple concatenation to align and fuse visual feature maps and linguistic feature vectors, respectively. For replacing simple concatenation, some prior works use cross-modal attention to focus on important pixel regions and informative keywords for long-range cross-modal context \cite{shi2018key,ye2019cross,chen2019see}. Besides, some other works use complex visual reasoning to capture more explainable cross-modal context \cite{huang2020referring,hui2020linguistic,yang2021bottom}.
Recently, Vision transformer~(ViT)~\cite{dosovitskiy2020image} has been proposed as a new visual network paradigm. Due to its compatibility with multi-modal data, some works use it to jointly encode visual and linguistic features for intensive cross-modal alignment~\cite{ding2021vision,li2021referring,wang2022cris,yang2022lavt}.

\section{Methods}

\subsection{Overall Pipeline}

To ensure the generality of our framework, the WiCo is designed to be compatible with arbitrary top-down and bottom-up methods. As shown in Figure~{\color{red}{\ref{fig:framework}}}, WiCo has three parts: top-down branch, bottom-up branch and ``Interaction then Integration". Top-down branch is used to deploy top-down methods. Bottom-up branch is used to equip bottom-up methods. ``Interaction then Integration" is the key component of WiCo which is used to build cooperation between top-down and bottom-up branches for achieving a win-win improvement.

Top-down style methods are essentially a cross-modal match network~\cite{yu2018mattnet}. It uses the pretrained detector and cross-modal match network to obtain instance masks $ \mathcal{M} = \{m^1 \in \{0, 1\}^{{H}\times{W}}, m^2, ..., m^n\} $, cross-modal instance embeddings $ \mathcal{E} = \{E^1_i \in \mathbb{R}^{C}, E^2_i, ..., E^n_i\} $ and cross-modal alignment scores $ \mathcal{S} = \{s^1, s^2, ..., s^n\} $. In general, top-down branch outputs a instance triplet set $\{\mathcal{M}, \mathcal{E}, \mathcal{S}\} = \{(m^1, E^1_i, s^1), (m^2, E^2_i, s^2), ..., (m^n, E^n_i, s^n)\}$. Extracting segmentation results $ P_{td} $ from triplet set is fomulated as:
\begin{equation}
\begin{split}
P_{td} = m^{\texttt{argmax}(\mathcal{S})} * s^{\texttt{argmax}(\mathcal{S})},
\end{split}
\end{equation}
where $ P_{td} $ is the segmentation logits results. The binary segmentation results are $ m^{\texttt{argmax}(\mathcal{S})} $.

Bottom-up methods are essentially a cross-modal fusion network~\cite{hu2016natural}. It uses a cross-modal fuse network to jointly encode images and texts to cross-modal pixel embeddings 
$ E_p \in \mathbb{R}^{C\times{H}\times{W}}$ and decode cross-modal pixel embeddings to segmentation results $ P_{bu} \in \mathbb{R}^{{H}\times{W}} $. Decoding cross-modal pixel embeddings into segmentation results is formulated as:
\begin{equation}
\begin{split}
P_{bu} = \sigma(\texttt{Linear}(E_p)),
\end{split}
\label{pred_bu}
\end{equation}
where $ \texttt{Linear}(\cdot) $ denotes 1x1 convolution for logit regression and $\sigma(\cdot)$ is sigmoid function. $ P_{bu} $ is the probability map and the binary segmentation results are extracted from it by threshold $ \tau $~($ P_{bu} > \tau $). In general, bottom-up branch outputs cross-modal pixel embeddings and segmentation results.

\begin{figure}[t]
\centering
\includegraphics[width=0.46\textwidth]{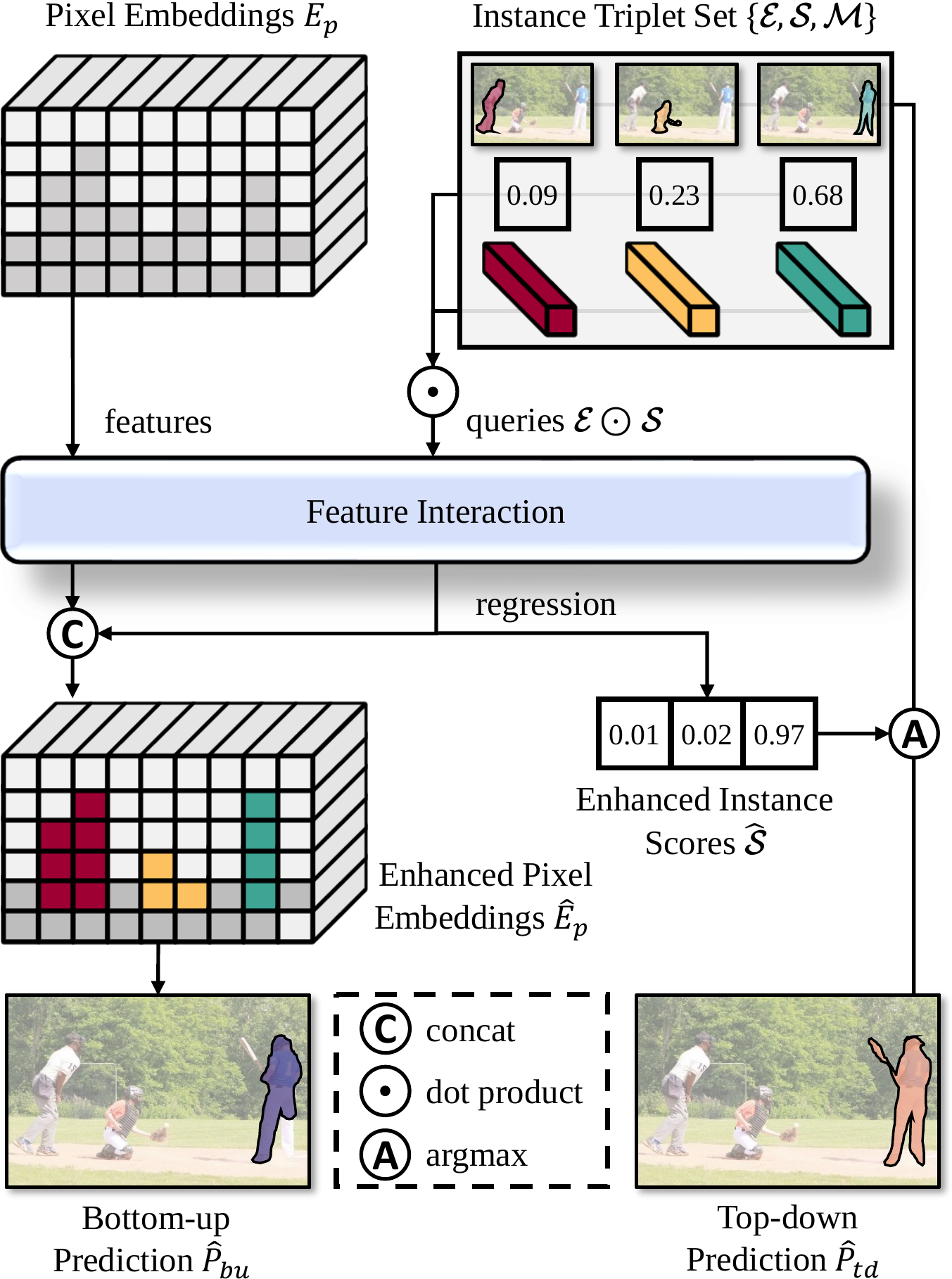} 
\caption{\textbf{Complementary Feature Interaction.} 
The modulated instance embeddings $\mathcal{E} \odot \mathcal{S}$ and pixel embeddings $ E_p $ are input into the ``Feature Interaction" for generating enhanced instance embeddings $\hat{\mathcal{E}}$. Enhanced instance embeddings are used to predict enhanced alignment scores $\hat{\mathcal{S}}$ for generating new top-down segmentation results $\hat{P}_{td}$. The enhance instance embeddings are also assigned to corresponding pixels of pixel embeddings for enhancing pixel embeddings $\hat{E}_p$ and generating new bottom-up segmentation $\hat{P}_{bu}$.}
\label{fig:cfi}
\end{figure}

``Interaction then Integration" is designed to exploit the complementary nature of top-down and bottom-up methods. To complement on interaction aspect, the outputs of bottom-up branch and top-down branch are input into CFI~(Section~{\color{red}{\ref{subsec:cfi}}}) for updating features and results. To complement on integration aspect, the updated results are input into GSI~(Section~{\color{red}{\ref{subsec:gsi}}}) to integrate results.

\subsection{Complementary Feature Interaction}
\label{subsec:cfi}

The detailed calculation flow is illustrated in Figure~{\color{red}{\ref{fig:cfi}}}.
Suppose that we already acquire pixel embeddings $ E_p $ from bottom-up branch and instance triplet set $\{\mathcal{M}, \mathcal{E}, \mathcal{S}\}$ from top-down branch, we hope that CFI can let the fine-grained information of pixel embeddings and object-centric information of instance triplet set enhance each other.

\paragraph{Top-down for Bottom-up.}
For enhancing pixel embeddings $ E_p $, we assign object-centric information of each enhanced instance embeddings $ \hat{\mathcal{E}} $ to corresponding pixels according to the instance masks $ \mathcal{M} $ and concatenate these instance embeddings with raw pixel embeddings to generate enhanced pixel embeddings $ \hat{E}_{p} $:
\begin{equation}
\begin{split}
\hat{E}_{p}^{\{x,y\}} = \texttt{concat}(E_{p}^{\{x,y\}} ; \sum^{n}_{j=1} \mathds{1}_{\{m^{j}[x, y] = 1\}}\hat{E}_i^{j}),
\end{split}
\end{equation}
where $ E_{p}^{\{x,y\}} $ denotes enhanced pixel embeddings at $ (x, y) $ pixel location and $ E_{i}^{j} $ is the enhanced instance embeddings of $ i $-th instance. $ \mathds{1}_{\{m^{j}[x, y] = 1\}} $ is the indicator function, where it is equal to 1 when the $(x,y)$ pixel location of $j$-th mask is 1, and 0 otherwise. The enhanced pixel embeddings are then decoded to new bottom-up results:
\begin{equation}
\begin{split}
\hat{P}_{bu} = \texttt{sigmoid}(\texttt{Linear}(\hat{E}_p)),
\end{split}
\end{equation}
where the $ \texttt{Linear}(\cdot) $ shares same weights with Eq.~{\color{red}{\ref{pred_bu}}}.

\paragraph{Bottom-up for Top-down.} 
For enhancing instance embeddings $\mathcal{E}$, we use vision transformer decoder~\cite{cheng2021masked} as ``Feature Interaction" module to refine the instance embeddings by fine-grained information of pixel embeddings $ E_p $. Before inputting, the instance embeddings are modulated by cross-modal alignment scores $ \mathcal{S} $ for preserving cross-modal information:
\begin{equation}
\begin{split}
\mathcal{E} \odot \mathcal{S} = \{E_p^1 * s^1, E_p^2 * s^2, ..., E_p^n * s^n\}.
\end{split}
\end{equation}
Then, transformer decoder sets these modulated instance embeddings $ \mathcal{E} \odot \mathcal{S} $ as queries to generate enhanced instance embeddings $ \hat{\mathcal{E}} $ and predict enhanced alignment scores $ \hat{\mathcal{S}} $. With new alignment scores, we can update the segmentation results of top-down branch:
\begin{equation}
\begin{split}
\hat{P}_{td} = m^{\texttt{argmax}(\hat{\mathcal{S}})} * \hat{s}^{\texttt{argmax}(\hat{\mathcal{S}})}.
\end{split}
\end{equation}

\subsection{Gaussian Scoring Integration}
\label{subsec:gsi}

After obtaining top-down results $ \hat{P}_{td} $ and bottom-up results $ \hat{P}_{bu} $, we use GSI to integrate them for generating more robust and higher-performance results. GSI has three steps: Distribution Prediction, Score Sampling and Results Blend. The details of three steps are introduced below:

\paragraph{Distribution Prediction.}
Because of the uncertainty, we set the performance score as a latent variable following a specific distribution. Due to the excellent computability, we use gaussian distribution to model the performance distribution~\cite{kingma2013auto}. For representing gaussian distribution, we predict the mean $ \mu $ and standard deviation $ \sigma $ according to the results and features of two branches:
\begin{align}
\mu_{td}, \sigma_{td} &= \texttt{split}(\texttt{MLP}(\hat{E}_i^{\texttt{argmax}(\hat{\mathcal{S}})}), \\
\mu_{bu}, \sigma_{bu} &= \texttt{split}(\texttt{MLP}(\texttt{GAP}(E_p \odot \hat{P}_{bu}))),
\end{align}
where $ \texttt{MLP}(\cdot) $ denotes 3 fully connected layers, $ \texttt{GAP}(\cdot) $ denotes global average pooling operation and $ \texttt{split}(\cdot) $ denotes channel split operation. With predicted mean and standard deviation, we obtain the performance distribution of bottom-up and top-down branches, i.e., $ \mathcal{N}(\mu_{bu}, \sigma_{bu}) $ and $ \mathcal{N}(\mu_{td}, \sigma_{td}) $.

\begin{table*}[t]
\centering                         
\renewcommand{\arraystretch}{1.3}  
\setlength{\tabcolsep}{2mm}        
\footnotesize                      
\begin{tabular}{r|c|lll|lll|l}
\noalign{\hrule height 1.5pt}
\multicolumn{1}{c|}{\multirow{2}{*}{Method}} & \multirow{2}{*}{Type} & \multicolumn{3}{c|}{RefCOCO} & \multicolumn{3}{c|}{RefCOCO+} & \multicolumn{1}{c}{RefCOCOg} \\
\cline{3-9}
& & val & test A & test B & val & test A & test B & val   \\
\hline
MAttNet~\pub{\cite{yu2018mattnet}} & TD & 56.51 & 62.37 & 51.70 & 46.67 & 52.39 & 40.08 & -     \\
NMTree~\pub{\cite{liu2019learning}} & TD &  56.59 & 63.02 & 52.06 & 47.40 & 53.01 & 41.56 & -     \\
CAC~\pub{\cite{chen2019referring}}  & TD &  58.90 & 61.77 & 53.81 & - & - & - & 44.32 \\
\hline
MCN~\pub{\cite{luo2020multi}}  & BU & 62.44 & 64.20 & 59.71 & 50.62 & 54.99 & 44.69 & -     \\
CMPC~\pub{\cite{huang2020referring}}  & BU & 61.36 & 64.53 & 59.64 & 49.56 & 53.44 & 43.23 & 39.98 \\
LSCM~\pub{\cite{hui2020linguistic}}   & BU & 61.47 & 64.99 & 59.55 & 49.34 & 53.12 & 43.50 & 48.05 \\
CGAN~\pub{\cite{luo2020cascade}}  & BU & 64.86 & 68.04 & 62.07 & 51.03 & 55.51 & 44.06 & 46.54 \\
BUSNet~\pub{\cite{yang2021bottom}}    & BU & 62.56 & 65.61 & 60.38 & 50.98 & 56.14 & 43.51 & 49.98 \\
EFN~\pub{\cite{feng2021encoder}}  & BU & 62.76 & 65.69 & 59.67 & 51.50 & 55.24 & 43.01 & - \\
LTS~\pub{\cite{jing2021locate}}   & BU & 65.43 & 67.76 & 63.08 & 54.21 & 58.32 & 48.02 &  - \\
VLT~\pub{\cite{ding2021vision}}   & BU & 65.65 & 68.29 & 62.73 & 55.50 & 59.20 & 49.36 & 49.76 \\
ResTR~\pub{\cite{kim2022restr}}   & BU & 67.22 & 69.30 & 64.45 & 55.78 & 60.44 & 48.27 & - \\
CRIS~\pub{\cite{wang2022cris}}    & BU & 69.52 & 72.72 & 64.70 & 61.39 & 67.10 & 52.48 & 55.77$^{*}$ \\
LAVT~\pub{\cite{yang2022lavt}}    & BU & 72.73 & 75.82 & 68.79 & 62.14 & 68.38 & 55.10 & 60.50 \\
SeqTR~\pub{\cite{zhu2022seqtr}}   & BU & 67.26 & 69.79 & 64.12 & 54.14 & 58.93 & 48.19 & - \\
CoupleAlign~\pub{\cite{zhang2022coupalign}} & BU & 74.70 & 77.76 & 70.58 & 62.92 & 68.34 & 56.69 & - \\
\hline
\rowcolor{aliceblue!60} WiCo~\bsl{VLT + MAttNet} & TD+BU & 71.74\increase{6.09} & 74.07\increase{5.78} & 67.23\increase{4.50} & 60.17\increase{4.67} & 65.15\increase{5.95} & 53.55\increase{4.19} & 53.37\increase{3.61} \\
\rowcolor{aliceblue!60} WiCo~\bsl{CRIS + MAttNet} & TD+BU & 73.46\increase{3.94} & 76.95\increase{4.23} & 68.08\increase{3.38} & 63.42\increase{2.03} & 69.17\increase{2.07} & 55.76\increase{3.28} & 60.17\increase{4.4} \\
\rowcolor{aliceblue!60} WiCo~\bsl{LAVT + MAttNet} & TD+BU & \textbf{75.50}\increase{6.66} & \textbf{78.07}\increase{2.25} & \textbf{71.30}\increase{2.51} & \textbf{65.75}\increase{3.61} & \textbf{70.52}\increase{2.14} & \textbf{57.14}\increase{2.04} & \textbf{61.27}\increase{0.77} \\
\noalign{\hrule height 1.5pt}
\end{tabular}
\caption{\textbf{Main results} on three classical datasets (RefCOCO, RefCOCO+ and RefCOCOg). "TD" denotes top-down methods. "BU" denotes bottom-up methods. The \textbf{\color{purple}{improvement}} is calculated based on bottom-up method. $^{*}$ denotes the results are re-implemented by us.
}
\label{tab:main_res}
\end{table*}

\paragraph{Score Sampling.}
Performance distribution indicates the confidence score range of prediction. We sample a value from the performance distribution as the detailed confidence score of this prediction.
For differentiable optimization, we utilize re-parameterization trick~\cite{kingma2013auto} to modify the sampling process:
\begin{align}
\begin{split}
\mathrm{IoU}_{td} &= \mu_{td} + \sigma_{td} * \epsilon, \epsilon\sim\mathcal{N}(0, \textbf{I}), \\
\mathrm{IoU}_{bu} &= \mu_{bu} + \sigma_{bu} * \epsilon, \epsilon\sim\mathcal{N}(0, \textbf{I}), 
\end{split}
\end{align}
where $\mathrm{IoU}_{td}$ and $\mathrm{IoU}_{bu}$ denote confidence score of top-down and bottom-up branch results. For optimizing the distribution prediction model, we calculate smooth-l1 loss between predicted confidence score and ground truth IoU value.

\paragraph{Results Blend.}
Note that $ \texttt{argmax}(\cdot) $ is essentially a non-differentiable operation during gradient backward, we adopt a differentiable implementation~\cite{NIPS2017_7a98af17} of $ \texttt{argmax}(\cdot) $ function during training pharse:
\begin{align}
\lambda = \texttt{one-hot}(\texttt{argmax}(\mathcal{\hat{S}})) + \hat{\mathcal{S}} - \texttt{sg}(\hat{\mathcal{S}}),
\end{align}
where $ \lambda \in \{0, 1\}^{n} $ is a binary vector to indicate the index of the max value, $ \texttt{one-hot}(\cdot) $ is the one-hot encoding function and $ \texttt{sg}(\cdot) $ is the stop gradient operation. The $ \lambda $ is used to build differentiable segmentation results of top-down branch $ \hat{P}^{'}_{td} $:
\begin{align}
\hat{P}^{'}_{td} = \sum^{n}_{j=1} m^{j} * \lambda^{j} * s^{j},
\end{align}
where $n$ is the number of instances. For generating final segmentation results, we use confidence scores to calculate a weighted sum results of top-down and bottom-up branches:
\begin{equation}
\begin{split}
\hat{P} = (\hat{P}^{'}_{td} * \mathrm{IoU}_{td} + \hat{P}_{bu} * \mathrm{IoU}_{bu})/2.
\end{split}
\end{equation}
The final results $ \hat{P} $ are used to calculate segmentation loss with ground truth mask during training phrase and are decoded to binary mask by threshold $ \tau $ during inference phrase.

\begin{figure*}[t]
\centering
\subfigtopskip=2pt    
\subfigbottomskip=1pt 
\subfigcapskip=-2pt   
\begin{minipage}[b]{0.24\linewidth}
\subfigure[GSI (Top-down)]{
    \includegraphics[width=0.98\linewidth]{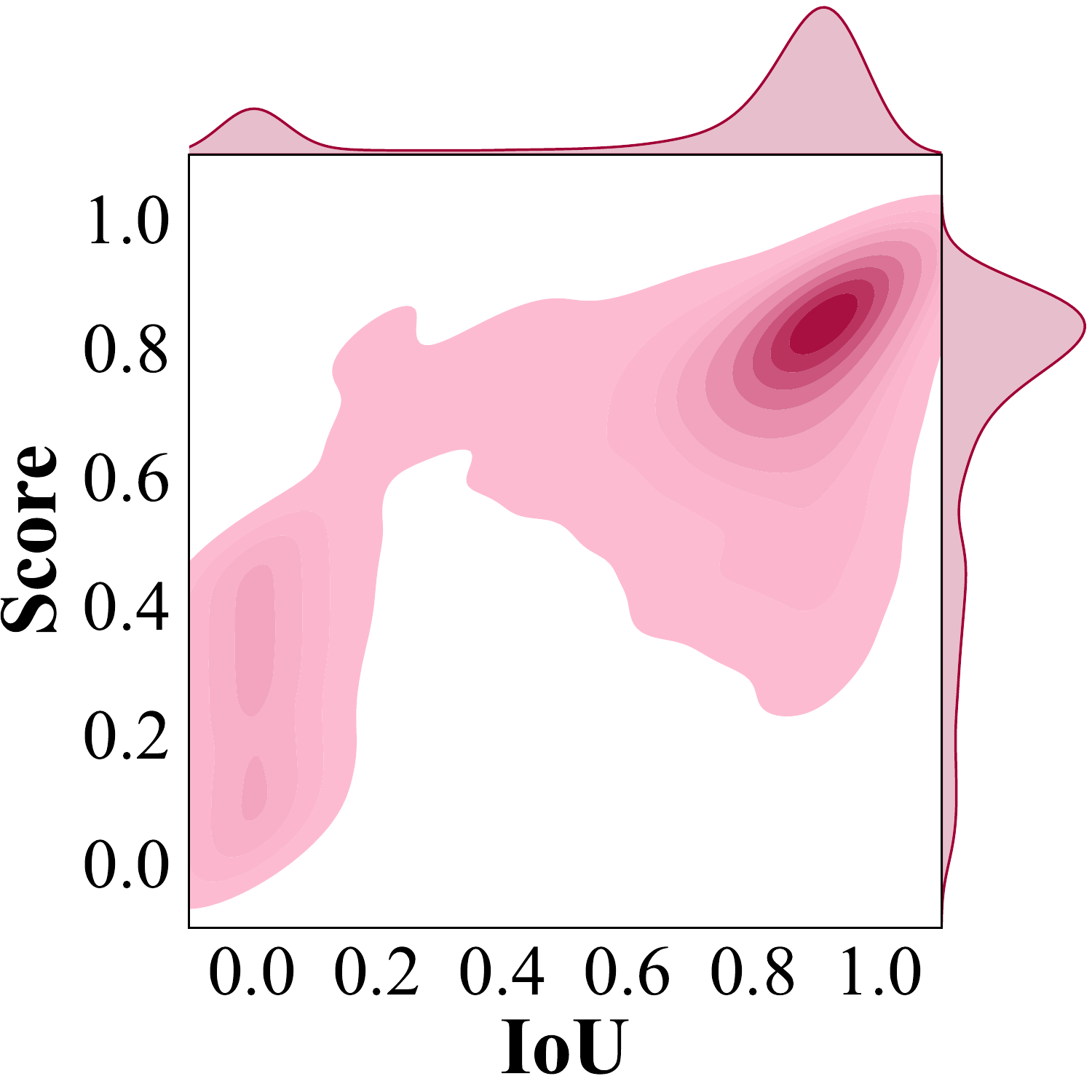}
}
\end{minipage}
\hfill
\begin{minipage}[b]{0.24\linewidth}
\subfigure[SI (Top-down)]{
    \includegraphics[width=0.98\linewidth]{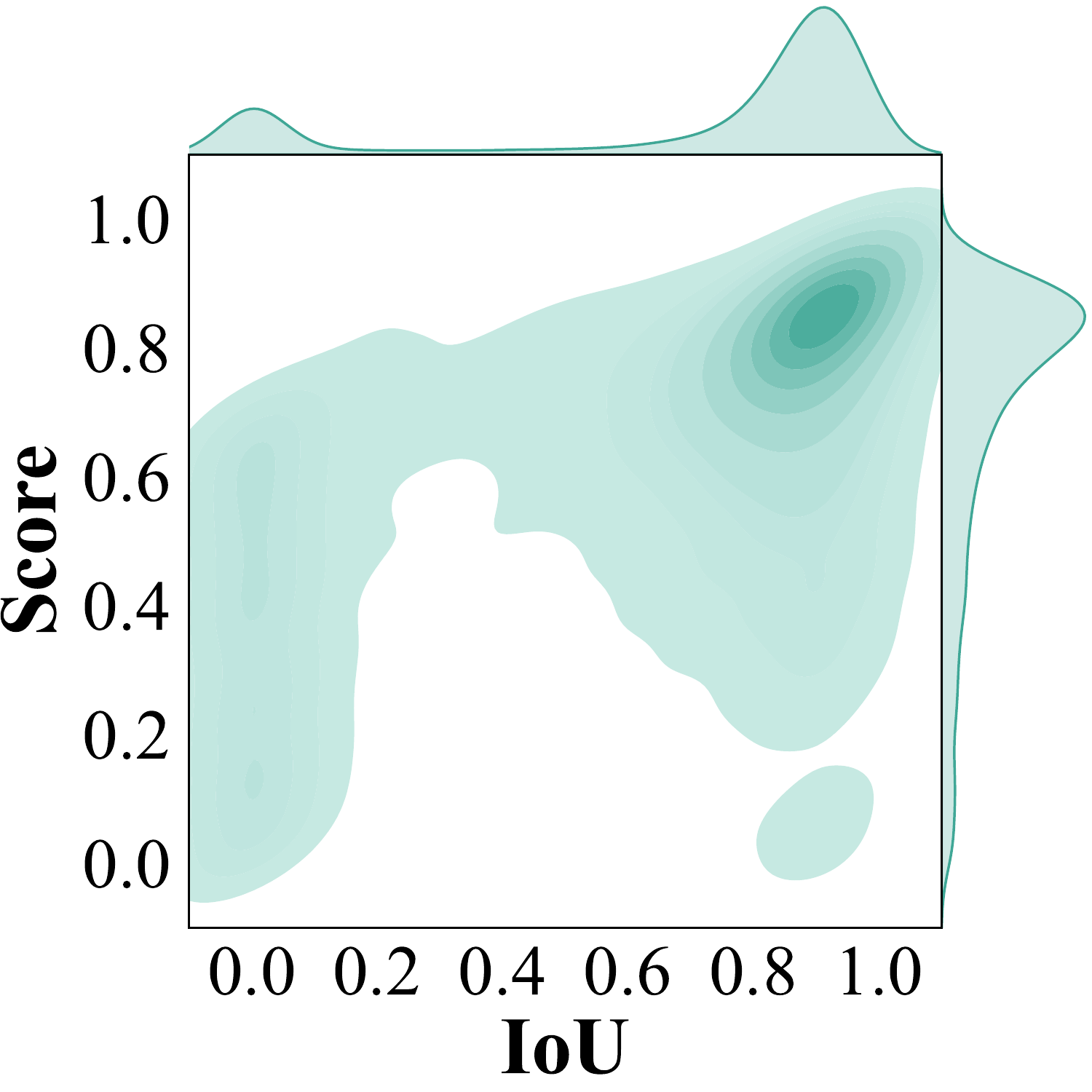}
}
\end{minipage}
\hfill
\begin{minipage}[b]{0.24\linewidth}
\subfigure[GSI (Bottom-up)]{
    \includegraphics[width=0.98\linewidth]{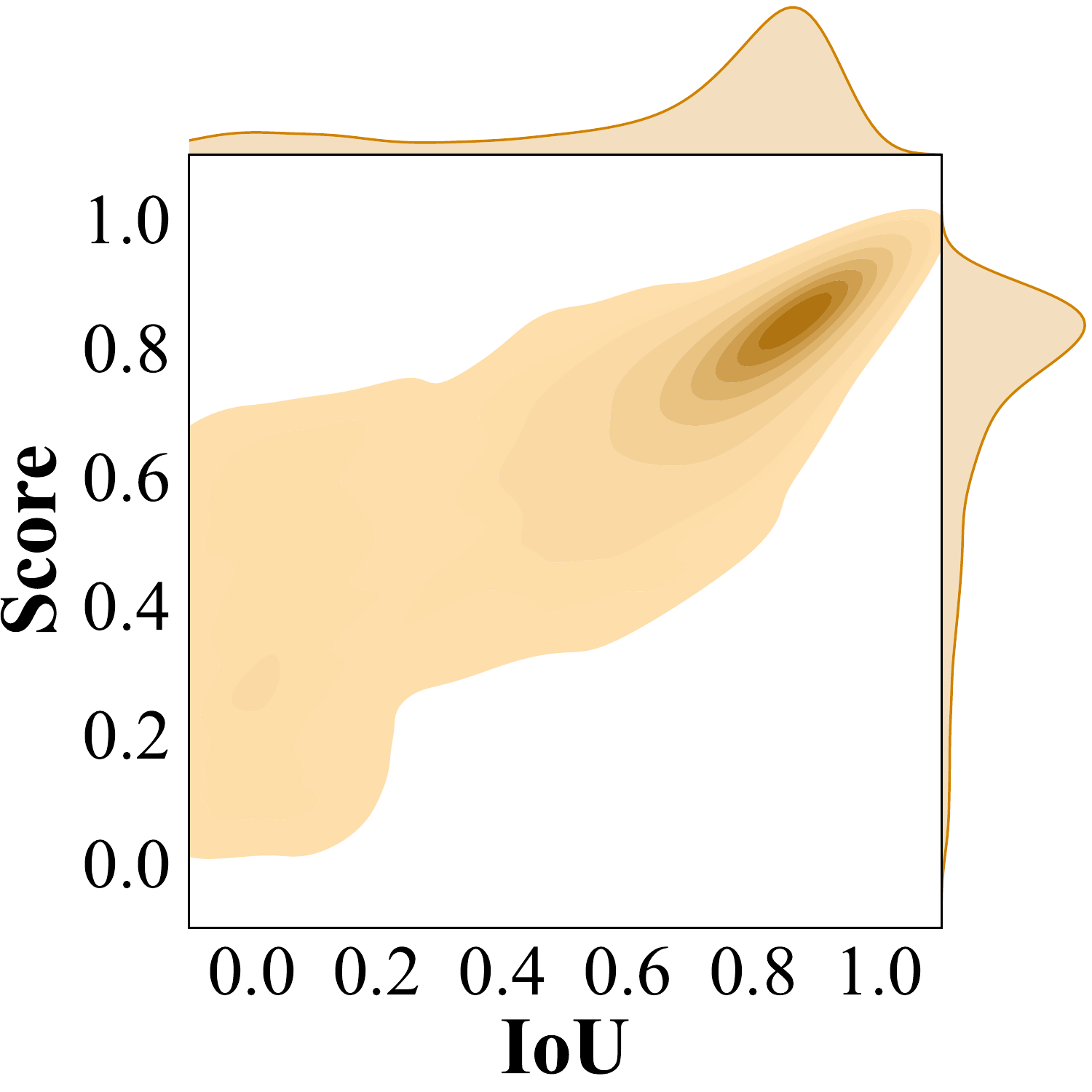}
}
\end{minipage}
\hfill
\begin{minipage}[b]{0.24\linewidth}
\subfigure[SI (Bottom-up)]{
    \includegraphics[width=0.98\linewidth]{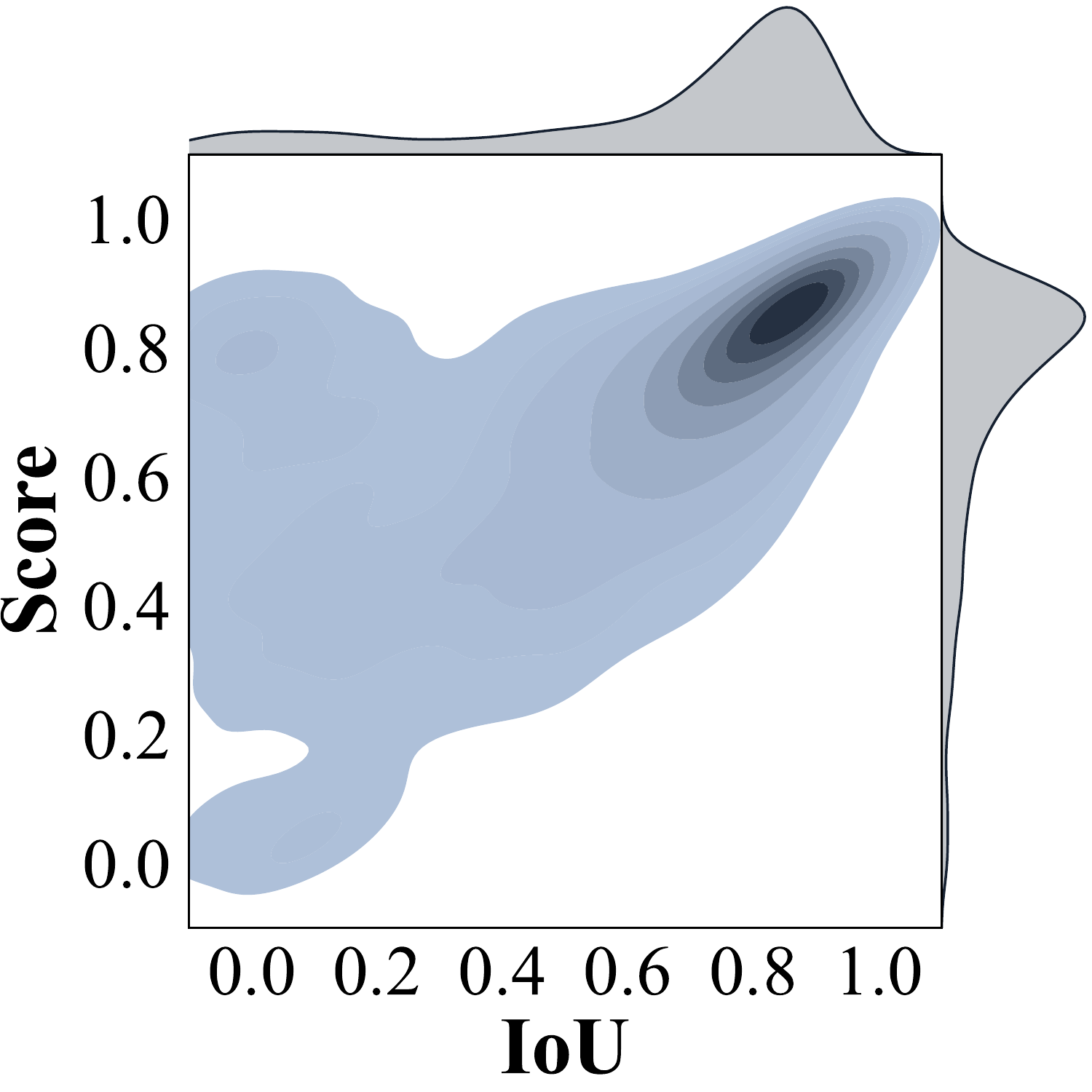}
}
\end{minipage} 
\caption{\textbf{Correlation between predicted confidence score and IoU.} The density map of samples from GSI and SI. Darker area indicates more samples are of the corresponding IoU (\%) value and confidence score. ``SI" is Scoring Integration, abbreviated from GSI by removing the gaussian distribution-based performance modeling. Marginal plots denote the distribution of confidence score and IoU.}
\label{fig:gsi_qual}
\end{figure*}

\section{Experiments}

\subsection{Experimental Setup}

Our model is evaluated on three standard referring image segmentation datasets: RefCOCO~\cite{yu2016modeling}, RefCOCO+~\cite{yu2016modeling} and RefCOCOg~\cite{mao2016generation}. 
For top-down branch, MAttNet~\cite{yu2018mattnet} is selected as the main equipment due to its simple structure and effectiveness. As for the bottom-up branch, several advanced and representative methods are selected, e.g., VLT~\cite{ding2021vision}, CRIS~\cite{wang2022cris} and LAVT~\cite{yang2022lavt}, to show the effectiveness and generality of our method.
The data preprocessing operations are in line with the original implementation of those selected methods.
Because MAttNet is an early method that has an obsolete instance extractor, Mask2Former~\cite{cheng2021masked} (ResNet-50) is adopted as an instance extractor to compensate for the top-down branch to avoid the cask effect, which improves the performance of MAttNet from 56.51 to 62.62 on RefCOCO val set. 
Based on previous works \cite{luo2020multi,ding2021vision}, mask IoU is adopted to evaluate the performance of methods.
To reduce the training cost, the selected models are initialized by pretrained weights and just finetune when inserting them into our framework. 
AdamW \cite{loshchilov2017decoupled} is adopted as our optimizer, and the learning rate and weight decay are set to 1e-5 and 5e-2. We train our models for 5,000 iterations on an NVIDIA V100 with a batch size of 24.
To binarize the probability map and get segmentation results, the threshold $ \tau $ is set to 0.35 to calibrate previous works~\cite{ding2021vision}.

\subsection{Quantitative Analysis}

\paragraph{Main Results.}
Table~{\color{red}{\ref{tab:main_res}}} reports the comparison results between our method and previous state-of-the-art methods in three common datasets, i.e., RefCOCO, RefCOCO+ and RefCOCOg. 
Some top-down and bottom-up methods that are easy to reproduce are selected for benchmark. Specifically, there are three combinations, i.e., VLT + MAttNet, CRIS + MAttNet and LAVT + MAttNet. Because bottom-up methods are mainstream methods, we mainly describe the performance improvement based on bottom-up methods in Table~{\color{red}{\ref{tab:main_res}}}.
Utilizing WiCo to incorporate these three model combinations, the fusion results improve the results of VLT, CRIS and LAVT by 6.09\%, 3.94\% and 6.66\% on RefCOCO val split, 5.78\%, 4.23\%, 2.25\% on RefCOCO testA split and 4.5\%, 3.38\% and 2.51\% on RefCOCO testB split. Other datasets also consistently show the performance improvements of our method over the selected baseline models.

\begin{table}
\centering                          
\renewcommand{\arraystretch}{1.35}  
\setlength{\tabcolsep}{2mm}         
\footnotesize                       
\begin{tabular}{z{42}x{38}|y{34}y{34}|y{39}}
\noalign{\hrule height 1.5pt}
\multicolumn{1}{c}{Integration}  & Interaction & \multicolumn{1}{c}{$ \rm IoU_{\alpha} $}  & \multicolumn{1}{c|}{$ \rm IoU_{\beta} $}      & \multicolumn{1}{c}{$ \rm IoU_{+} $} \\
\hline
Intersection & -            & \multirow{3}*{65.65} & \multirow{3}*{62.62}       & 63.85{\decrease{4.70}} \\
Union        & -            &                      &                            & 67.79{\decrease{0.76}} \\
Average      & -            &                      &                            & 68.55 \\
\hline
\rowcolor{aliceblue!60} SI  & -   &                       &                            & 68.95{\increase{0.40}} \\
\rowcolor{aliceblue!60} GSI & -   & \multirow{-2}*{65.65} & \multirow{-2}*{62.62} & 69.63{\increase{1.08}} \\
\hdashline
\rowcolor{aliceblue!60} SI  & CFI &               &                   &  
70.51{\increase{1.96}} \\
\rowcolor{aliceblue!60} GSI & CFI &  \multirow{-2}*{68.07{\increase{2.42}}} & \multirow{-2}*{65.34{\increase{2.72}}}                 & \textbf{71.74}{\increase{3.19}} \\
\noalign{\hrule height 1.5pt}
\end{tabular}
\caption{\textbf{Diagnostic Experiments.} $ \rm IoU_{\alpha} $, $ \rm IoU_{\beta} $ and $ \rm IoU_{+} $ denotes the IoU of model $\alpha$~(VLT), model $\beta$~(MAttNet) and integration results, respectively. ``Intersection", ``Union" and ``Average" means taking the intersection, union and average of the top-down and bottom-up results as the fusion result. ``SI" is Scoring Integration, abbreviated from GSI by removing the gaussian distribution-based performance modeling. ``Average" scheme is set as the baseline for comparison.}
\vspace{-2.7pt}
\label{tab:main_abl}
\end{table}

\paragraph{Different Results Integration Strategies.} 
%
In Table~{\color{red}{\ref{tab:main_abl}}}, we attempt different results integration strategies and check if these integration strategies can boost the integration results of top-down and bottom-up branches.
In terms of results integration strategies, GSI is compared to three straight strategies, i.e., ``Intersection", ``Union", and ``Average". Although these strategies improve the performance of top-down and bottom-up methods, our proposed GSI still performs better than them, indicating that GSI provides a more perceptive and robust way to integrate results.
Moreover, we build an abbreviated version of GSI to check the effectiveness of the gaussian distribution-based performance modeling, i.e., Scoring Integration~(SI). Based on the experiment results that the GSI performs 0.68\% better than SI, it is concluded that the gaussian distribution-based performance modeling makes sense.

\paragraph{Effect of Feature Interaction.}
Feature interaction boosts results by improving the respective results of top-down and bottom-up branches.
For diagnosing if feature interaction is beneficial for final results, we conduct comparison experiments of WiCo with CFI and without CFI.
As shown in Table~{\color{red}{\ref{tab:main_abl}}}, WiCo with CFI improves baseline by 3.19\% and performs 2.11\% better than WiCo without CFI. 
The experiment results show that CFI effectively improves top-down and bottom-up branches by 2.42\% and 2.72\%. Moreover, it also shows that feature interaction~(CFI) boosts final performance on a different aspect than results integration~(GSI).

\begin{table}
\centering
\renewcommand{\arraystretch}{1.35}  
\setlength{\tabcolsep}{1.0mm}        
\footnotesize                      
\begin{tabular}{z{70}|x{53}z{38}|z{57}}
\noalign{\hrule height 1.5pt}
\multicolumn{1}{c|}{model~$ \alpha $+$ \beta $} & \multicolumn{1}{c}{$ \rm IoU_{\alpha} $+$ \rm IoU_{\beta} $} & \multicolumn{1}{c|}{$ \rm IoU_{+} $} & \multicolumn{1}{c}{Speed} \\
\hline
VLT$^\clubsuit$+CRIS$^\clubsuit$ & 65.65+69.52 & 70.15\increase{2.57} & 6.63 \pub{2.17+3.72+0.74} \\
VLT$^\clubsuit$+LAVT$^\clubsuit$ & 65.65+72.73 & 73.05\increase{3.86} & 10.2 \pub{2.17+7.27+0.74} \\
CRIS$^\clubsuit$+LAVT$^\clubsuit$ & 69.52+72.73 & 73.87\increase{2.75} & 11.7 \pub{3.72+7.27+0.74} \\
\hdashline
CAC{\color{red}{$^\varheartsuit$}}+MAttNet{\color{red}{$^\varheartsuit$}} & 63.43+62.62 & 62.72\decrease{0.31} & 4.43 \pub{2.11+1.88+0.44}  \\
\hline
\rowcolor{aliceblue!60} VLT$^\clubsuit$+MAttNet{\color{red}{$^\varheartsuit$}} & 65.65+62.62 & 69.63\increase{5.49} & 4.64 \pub{2.17+1.88+0.59}  \\
\rowcolor{aliceblue!60} CRIS$^\clubsuit$+MAttNet{\color{red}{$^\varheartsuit$}} & 69.52+62.62 & 73.01\increase{6.94} & 6.19 \pub{3.72+1.88+0.59}  \\
\rowcolor{aliceblue!60} LAVT$^\clubsuit$+MAttNet{\color{red}{$^\varheartsuit$}} & 72.73+62.62 & \textbf{74.33}\increase{6.66} & 9.74 \pub{7.27+1.88+0.59} \\
\noalign{\hrule height 1.5pt}
\end{tabular}
\caption{\textbf{The performance of different model combinations.} For checking the complementary effect between different models, model $\alpha$ and model $\beta$ are integrated by only GSI. $^\clubsuit$ and {\color{red}{$^\varheartsuit$}} denote bottom-up style methods and top-down style methods, respectively. Inference speed is acquired by counting the inference seconds of 100 samples. The \textbf{\color{purple}{increase}} value and \textbf{\color{gray!48}{decrease}} value are calculated by subtracting $\rm (IoU_{\alpha} + IoU_{\beta}) / 2 $ from $\rm IoU_{+} $, i.e., $\rm IoU_{+} - (IoU_{\alpha} + IoU_{\beta}) / 2 $.}
\label{tab:diff_model_combs}
\end{table}

\begin{figure*}[t]
\centering
\includegraphics[width=1.0\textwidth]{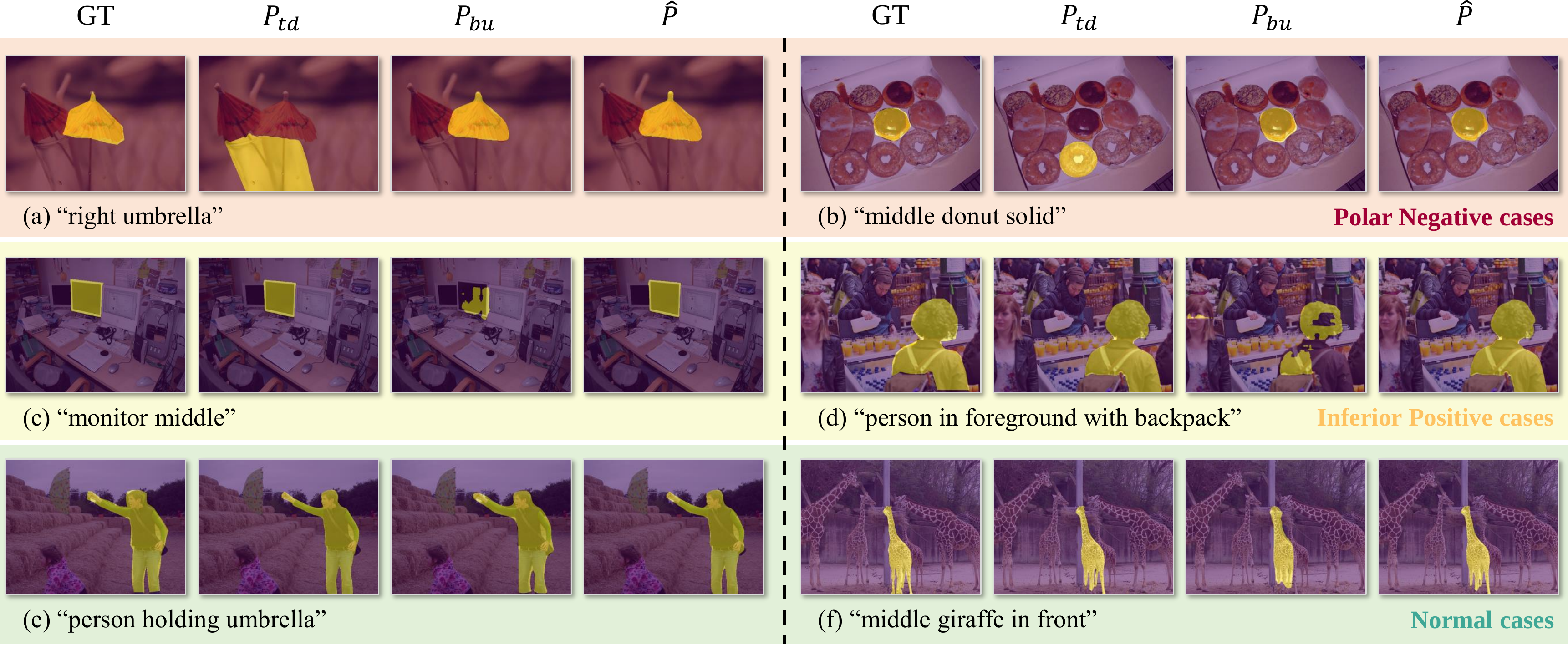}
\caption{\textbf{Qualitative segmentation reults of different cases.} ``GT", $P_{td}$, $P_{bu}$ and $\hat{P}$ denotes ground truth, original results of bottom-up branch, original results of top-down branch and the integration results of two branches. There are totally three types of cases selected for showing the effectiveness of our WiCo The first, second and third rows are polar negative cases, inferior positive cases and normal cases.}
\label{fig:main_qual}
\end{figure*}
\begin{figure}[t]
\centering
\includegraphics[width=1.0\linewidth]{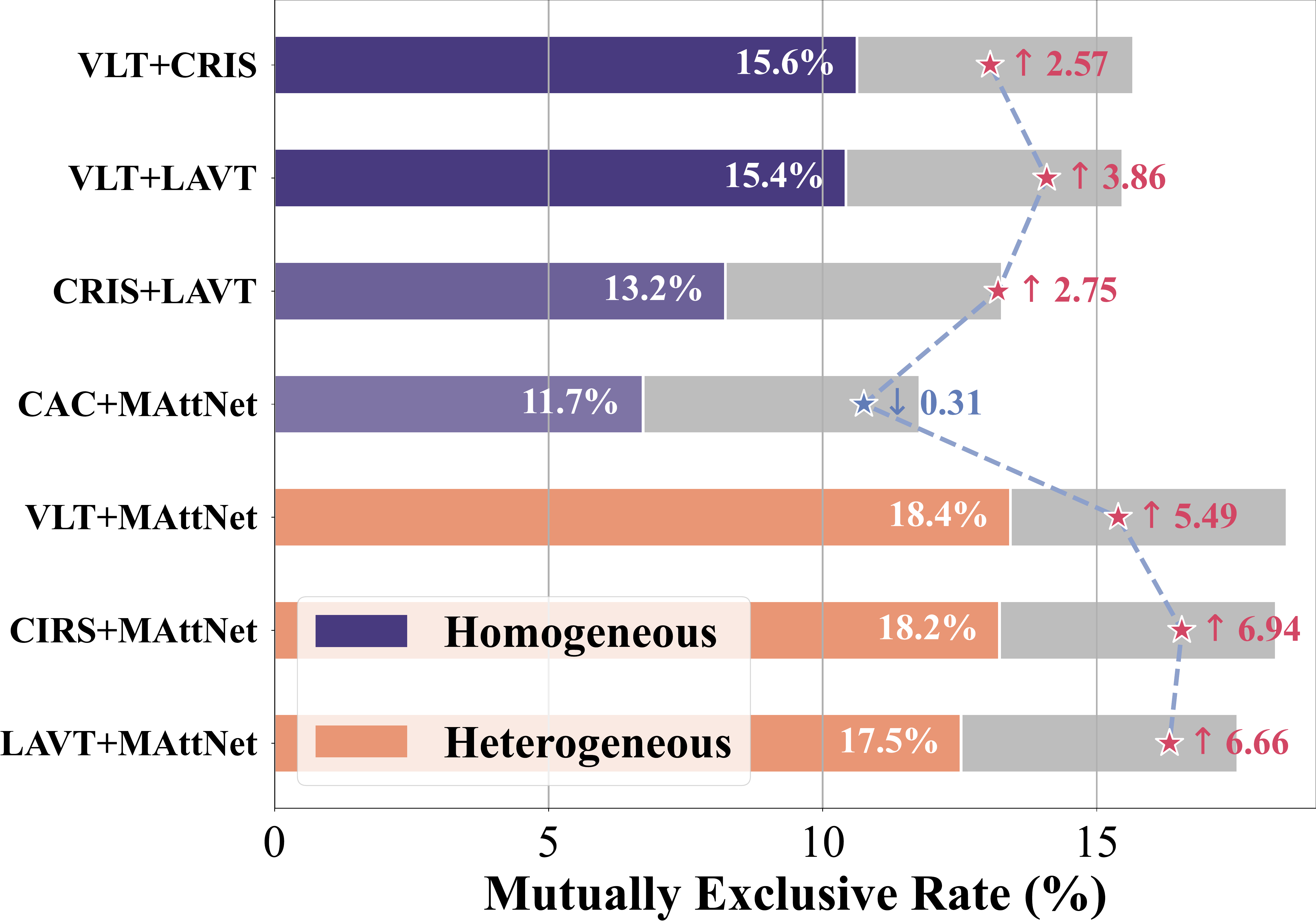} 
\caption{\textbf{The quantified analysis of complementary effect.} This figure is corresponding to Table~{\color{red}{\ref{tab:diff_model_combs}}}. \textbf{Mutually Exclusive Rate}~(MER) denotes the rate of samples in which only a single branch outputs positive prediction~($\mathrm{IoU} > 0.5$). A higher MER denotes the outputs of top-down and bottom-up branches are more complementary.}
\label{fig:bu_td_mer}
\end{figure}

\paragraph{Complementary Effect of Different Model Combinations.}
Three kinds of combinations are constructed (bottom-up + bottom-up, top-down + top-down and bottom-up + top-down) to check the complementary effect of different model combinations in Table~{\color{red}\ref{tab:diff_model_combs}}.
The model combination with two same kinds of models is defined as ``\textbf{Homogeneous}" combination. On the contrary, the model combination with two different kinds of models is defined as ``\textbf{Heterogeneous}" combination.
As shown in Table~{\color{red}\ref{tab:diff_model_combs}}, the experimental results can be split into three parts: bottom-up homogeneous combinations~(VLT+CRIS, VLT+LAVT and CRIS+LAVT), top-down homogeneous combinations~(MAttNet+CAC), and heterogeneous combinations between bottom-up and top-down methods~(VLT+MAttNet, CRIS+MAttNet and LAVT+MAttNet). Three bottom-up homogeneous combinations only improve original models by 2.57\%, 3.86\%, 2.75\% and the top-down homogeneous combination even degrade origin models by 0.31\%. However, three heterogeneous combinations consistently improve original models by a clear margin~(5.49\%, 6.94\%, 6.66\%). 
These results indicate that heterogeneous combinations have a stronger complementary effect than homogeneous combinations for boosting performance.
In order to quantify the complementary effect, ``\textbf{Mutually Exclusive Rate}"~(MER) is defined as a metric for analyzing. MER denotes the rate of samples in which only one of the top-down and bottom-up branches outputs positive prediction~($\mathrm{IoU} > 0.5$).
In Figure~{\color{red}{\ref{fig:bu_td_mer}}}, the MER of heterogeneous combinations is significantly higher than homogeneous combinations. 
These statistics results explain why the performance improvement of heterogeneous combinations is also remarkably higher than homogeneous combinations.

\subsection{Qualitative Analysis}

The key reason why GSI and SI achieve better performance than straight schemes is that they can roughly estimate the confidence of outputs to adaptively integrate the results.
For checking the precision of confidence regression, we plot the correlation between the predicted confidence score and real evaluation score~(IoU) in Figure~{\color{red}{\ref{fig:gsi_qual}}}. 
Whether the regression target is the performance score of top-down or bottom-up results, the plot results show that the confidence score predicted by GSI presents a more linear correlation with the evaluation score than SI, which demonstrates the gaussian distribution-based performance modeling of GSI is significant.

Some representative samples of three cases~(polar negative cases, inferior positive cases and normal cases) are selected to justify the refinement for PN and IP errors.
In Figure~{\color{red}{\ref{fig:main_qual}}}, first and second rows clearly depict the integration results of WiCo fix the obvious errors of original top-down and bottom-up results, which demonstrates the effectiveness of our method.
In Figure~{\color{red}{\ref{fig:main_qual}}}, third row also shows that our method can adaptively fetch better segmentation results from two branches.

\section{Conclusion}
Existing top-down and bottom-up methods fail to handle PN and IP errors.
Nevertheless, top-down and bottom-up methods can complement each other's flaws for better processing PN and IP errors according to our analysis.
To fully exploit the complementary nature, we follow a ``Interaction then Integration" paradigm to build WiCo mechanism for achieving a win-win improvement.
Specifically, CFI is proposed to let the prior object information of top-down branch and fine-grained information of bottom-up branch interact with each other for feature enhancement.
GSI is designed to model the performance distributions of two branches for adaptively integrating results of two branches.
We select some prominent top-down and bottom-up methods to equip our WiCo for experiments. The experiments consistently show that our WiCo can improve both top-down and bottom-up methods by a clear margin, which demonstrates the effectiveness of our methods.

\section*{Acknowledgements} 
This work was supported in part by the National Key R\&D Program of China (No.~2022ZD0118201), the Natural Science Foundation of China (No.~61972217, 32071459, 62176249, 62006133, 62271465), the Natural Science Foundation of Guangdong Province in China (No.~2019B1515120049).

\clearpage

\bibliographystyle{named}
\bibliography{ijcai23}

\clearpage
\appendix

\begin{center}
    \Large \bf Appendix
\end{center}
\vspace{15pt}

In this appendix, we first introduce how the IoU distribution is calculated and how the WiCo improves Top-down and Bottom-up methods from the perspective of IoU distribution curve~(Section~{\color{red}{\ref{sup_sec:iou_dist_curve}}}). Then we report extra quantitative results to analyze how the different proposals influence the performance of Top-down methods and the effect of proposed WiCo~(Section~{\color{red}{\ref{sup_sec:extra_quan_res}}}). Finally we provide extra qualitative cases to show how Complementary Feature Interaction~(CFI) improves Top-down and Bottom-up branches~(Section~{\color{red}{\ref{sup_sec:extra_qual_res}}}).

\begin{figure}[h]
\centering
\subfigtopskip=2pt    
\subfigbottomskip=1pt 
\subfigcapskip=-2pt   
\begin{minipage}[b]{0.48\linewidth}
    \subfigure[
    \textbf{\color{convnext_yellow}{Top-down}} vs \textbf{\color{convnext_purple}{WiCo}}
    ]{
        \includegraphics[width=1\linewidth]{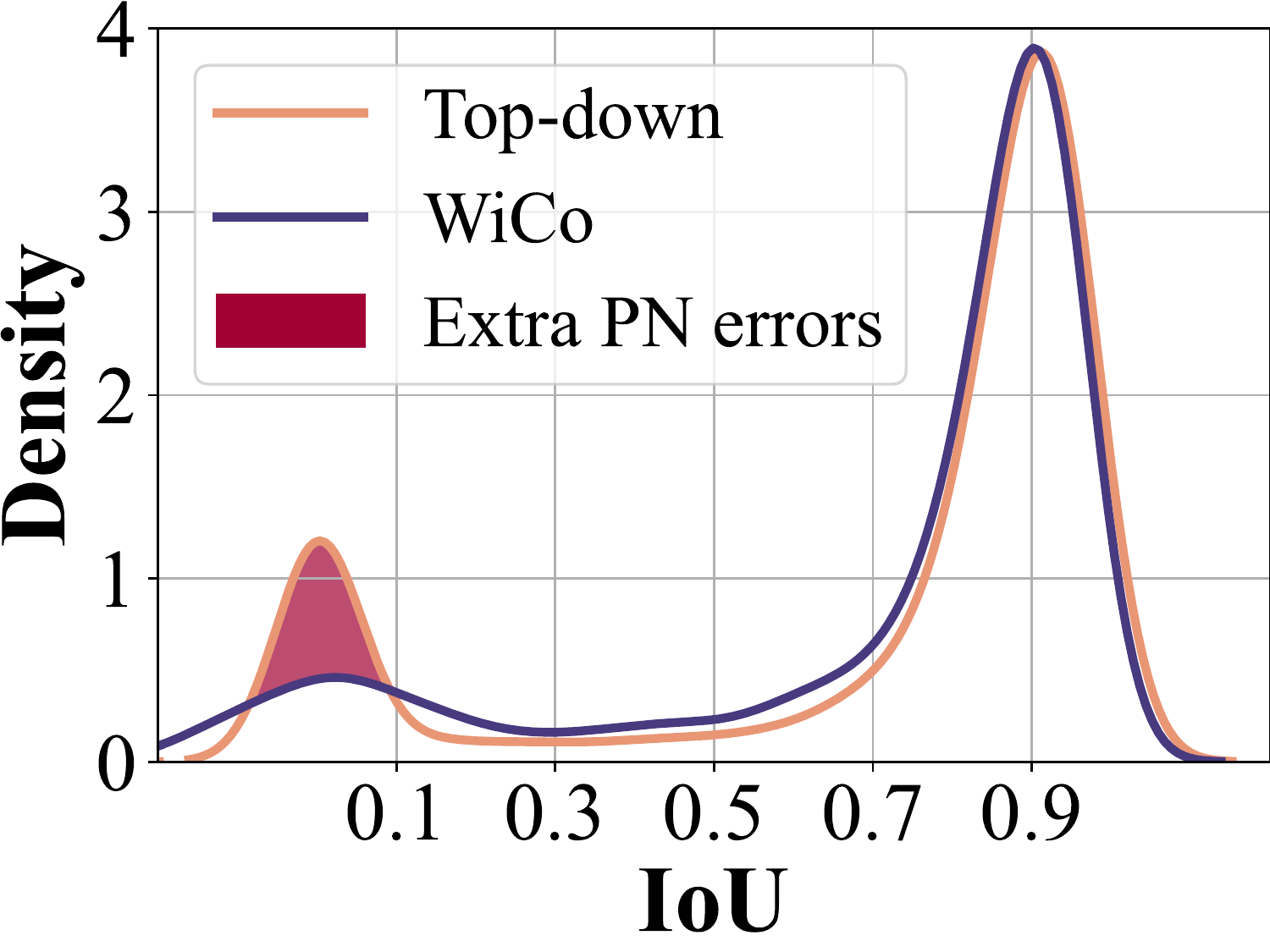}
    }
\end{minipage}
\hfill
\begin{minipage}[b]{0.48\linewidth}
    \subfigure[
    \textbf{\color{convnext_yellow}{Bottom-up}} vs \textbf{\color{convnext_purple}{WiCo}}
    ]{
        \includegraphics[width=1\linewidth]{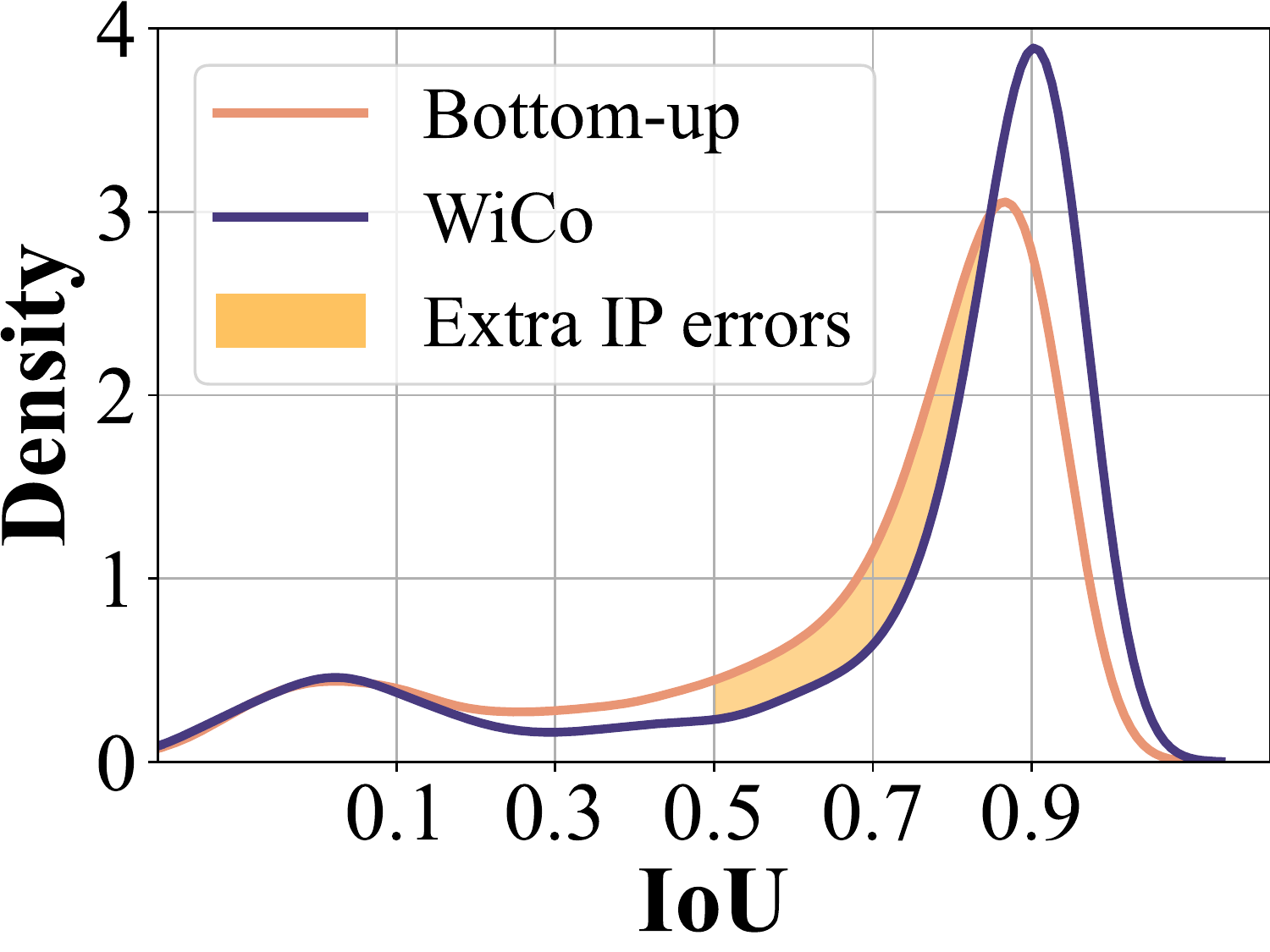}
    }
\end{minipage} 
\caption{\textbf{The IoU distribution comparison} between (a) Top-down method and WiCo, (b) Bottom-up method and WiCo. The extra \textbf{\color{cred}{Polar Negative}}~(PN) and \textbf{\color{cyellow}{Inferior Positive}}~(IP) errors for the top-down and bottom-up methods compared to WiCo are flagged.}
\vspace{-5pt}
\label{fig:sup_iou_dist_comp}
\end{figure}

\section{IoU Distribution Curve}
\label{sup_sec:iou_dist_curve}

\textbf{How does the IoU distribution calculate?} 
RefCOCO~\cite{yu2016modeling} val split is selected as our main dataset to plot the IoU distribution curve. The IoU distribution curve illustrates the probability density of samples with different IoU values. The calculation of probability density is mainly obtained by \textbf{kernel density estimation} which is the application of kernel smoothing for probability density estimation~\cite{rosenblatt1956remarks,parzen1962estimation}. We select Gaussian density function as the kernel function of kernel density estimation.

\noindent\textbf{How does the WiCo improve Top-down and Bottom-up methods?}
For detailedly checking how does WiCo improve Top-down and Bottom-up methods, we compare the IoU distribution of WiCo to Top-down method and Bottom-up method in Figure~{\color{red}{\ref{fig:sup_iou_dist_comp}}}. 
In Figure~{\color{red}{\ref{fig:sup_iou_dist_comp}}} (a), the positive set~(IoU$>$0.5) of IoU distribution of WiCo and Top-down method are nearly equivalent, while the negative set~(IoU$<$0.5) of IoU distribution of WiCo is flatter than Top-down method, especially those samples with nearly zero IoU~(IoU$\rightarrow$0), i.e., \textbf{Polar Negative}~(PN) errors.
In Figure~{\color{red}{\ref{fig:sup_iou_dist_comp}}} (b), the positive set of IoU distribution of WiCo is sharper than Bottom-up method so that it generates fewer low-quality positive samples~(IoU$\in[0.5,0.8]$), i.e., \textbf{Inferior Positive}~(IP) errors.
As the observations above, it is concluded that WiCo improves Top-down and Bottom-up methods by largely reducing PN and Inferior Positive~(IP) errors.

\begin{table}[t]
\centering
\small
\renewcommand{\arraystretch}{1.3}  
\setlength{\tabcolsep}{1.7mm}      
\footnotesize                      
\begin{tabular}{x{50}|z{25}|y{39}y{39}y{39}}
\noalign{\hrule height 1.5pt}
\multirow{2}{*}{Proposals} & \multicolumn{1}{c|}{\multirow{2}{*}{mAP}} & \multicolumn{3}{c}{RefCOCO} \\
\cline{3-5}
& & val & testA & testB \\
\hline
\multirow{2}{*}{Mask R-CNN}  & 27.88{\color{red}{$^\varheartsuit$}} & 56.51 & 62.37 & 51.70 \\
& 32.70$^\clubsuit$ & - & - & - \\
\hdashline
\rowcolor{aliceblue!60}
& 38.85{\color{red}{$^\varheartsuit$}} & \textbf{62.62}{\increase{6.11}} & \textbf{66.94}{\increase{4.57}} & \textbf{56.04}{\increase{4.34}} \\
\rowcolor{aliceblue!60}
\multirow{-2}{*}{Mask2former} & \textbf{43.69}$^\clubsuit$ & - & - & - \\ 
\noalign{\hrule height 1.5pt}
\end{tabular}
\caption{\textbf{The performance of MAttNet with different proposals}. Both Mask R-CNN and Mask2former set resnet50 as the backbone. {\color{red}{$^\varheartsuit$}} denotes the detector is trained on \textit{COCO} dataset without val/test sets of three classical datasets, i.e., \textit{RefCOCO}, \textit{RefCOCO+}, \textit{RefCOCOg}. $^\clubsuit$ denotes the detector is trained on full \textit{COCO} dataset.}
\vspace{-10pt}
\label{tab:MAttNet_w_diff_proposals}
\end{table}

\section{Extra Quantitative Results}
\label{sup_sec:extra_quan_res}

\textbf{Performance of different proposal extractor.} 
Works on Top-down style methods has been stalled for a long time so that the performance of top-down methods fall far behind bottom-up methods. 
In order to make top-down methods as the counterpart of advanced bottom-up methods, we first pretrain an advanced instance segmentation model~(Mask2former) as a proposal extractor for top-down methods.
The backbone network of instance segmentation model selects resnet50~\cite{he2016deep}. When training the Mask2former \cite{cheng2021masked}, we exclude the images contained in the val/test sets of \textit{RefCOCO}~\cite{yu2016modeling}, \textit{RefCOCO+}~\cite{yu2016modeling} and \textit{RefCOCOg}~\cite{mao2016generation} datasets. 
In order to check how these images influence the performance of the pretrain model, we report the comparison results of Mask2former trained on full \textit{COCO 2017} dataset and trained on \textit{COCO 2017} dataset without val/test sets of \textit{RefCOCO/RefCOCO+/RefCOCOg} datasets in Table~{\color{red}{\ref{tab:MAttNet_w_diff_proposals}}}. 
Besides, we also report the comparison results of Mask R-CNN \cite{he2017mask} (original proposal extractor of top-down methods) trained on full \textit{COCO 2014} dataset and trained on \textit{COCO 2014} dataset without val/test sets of \textit{RefCOCO/RefCOCO+/RefCOCOg} datasets in Table~{\color{red}{\ref{tab:MAttNet_w_diff_proposals}}}.

\noindent\textbf{Top-down method with different proposals.}
Different instance proposal extractors provide different quality proposals. 
We report the results of Top-down method~(MAttNet) with different instance proposal extractors in Table~{\color{red}{\ref{tab:MAttNet_w_diff_proposals}}}. 
With high-quality instance proposals of mask2former, the performance of MAttNet~\cite{yu2018mattnet} is improved by 6.11\%, 4.57\%, 4.34\% on \textit{RefCOCO} val, testA, testB split, which shows that the potential of Top-down method was heavily limited to antique instance proposal extractor in the past.
Moreover, MAttNet achieves comparable performance~(62.62 IoU) to bottom-up methods, like VLT~\cite{ding2021vision}~(65.65 IoU).

\begin{table}
\centering                         
\renewcommand{\arraystretch}{1.3}  
\setlength{\tabcolsep}{1.5mm}      
\footnotesize                      
\begin{tabular}{z{42}x{37}|y{39}y{39}|y{39}}
\noalign{\hrule height 1.5pt}
\multicolumn{1}{c}{Integration}  & Interaction & \multicolumn{1}{l}{$ \rm IoU_{\alpha} $}  & \multicolumn{1}{l|}{$ \rm IoU_{\beta} $}      & \multicolumn{1}{c}{$ \rm IoU_{+} $} \\
\hline
Intersection & -            & \multirow{3}*{65.65} & \multirow{3}*{56.51}       & 60.07{\decrease{6.06}} \\
Union        & -            &                      &                            & 65.38{\decrease{0.75}} \\
Average      & -            &                      &                            & 66.13 \\
\hline
\rowcolor{aliceblue!60} GSI & -   & 65.65 & 56.51 & 66.72{\increase{0.59}} \\
\rowcolor{aliceblue!60} GSI & CFI & 66.31{\increase{0.66}} & 59.32{\increase{2.81}} & \textbf{68.09}{\increase{1.96}}  \\
\noalign{\hrule height 1.5pt}
\end{tabular}
\caption{\textbf{Diagnostic Experiments.} $ \rm IoU_{\alpha} $, $ \rm IoU_{\beta} $ and $ \rm IoU_{+} $ denotes the IoU of model $\alpha$~(VLT), model $\beta$~(MAttNet) and fusion results. ``Intersection", ``Union" and ``Average" means taking the intersection, union and average of the top-down and bottom-up results as the fusion result. ``Average" scheme is set as the baseline for comparison. The proposals of MAttNet is from Mask R-CNN~(res50).}
\vspace{-5pt}
\label{tab:sup_main_abl}
\end{table}

\noindent\textbf{Dianoistic experiments with original proposals.}
For confronting the performance gap between Top-down methods and bottom-up methods, Top-down method are equipped with a high-performance instance proposal extractor~(mask2former).
To check if our method is still effective when the Top-down is equipped with a obsolete instance proposal extractor, we conduct diagnoistic experiments based on MAttNet with original proposal extractor.
In Table~{\color{red}{\ref{tab:sup_main_abl}}}, it shows that our proposed GSI improves baseline~(``Average" scheme) by 0.59\% and the integration results achieve 1.07\% and 10.21\% higher performance than Top-down and Bottom-up branches.
Besides, introducing our CFI to let Top-down and Bottom-up branches interact with each other still improves two-branches by 2.81\% and 0.66\%.
According to the experiment results, it is concluded that our proposed WiCo is still a highly effective cooperation scheme to exploit the complementary nature of two branches for improving both Top-down and Bottom-up methods even if Top-down branch adopts obsolete proposal extractor. 

\begin{figure}[t]
\centering
\includegraphics[width=1.0\linewidth]{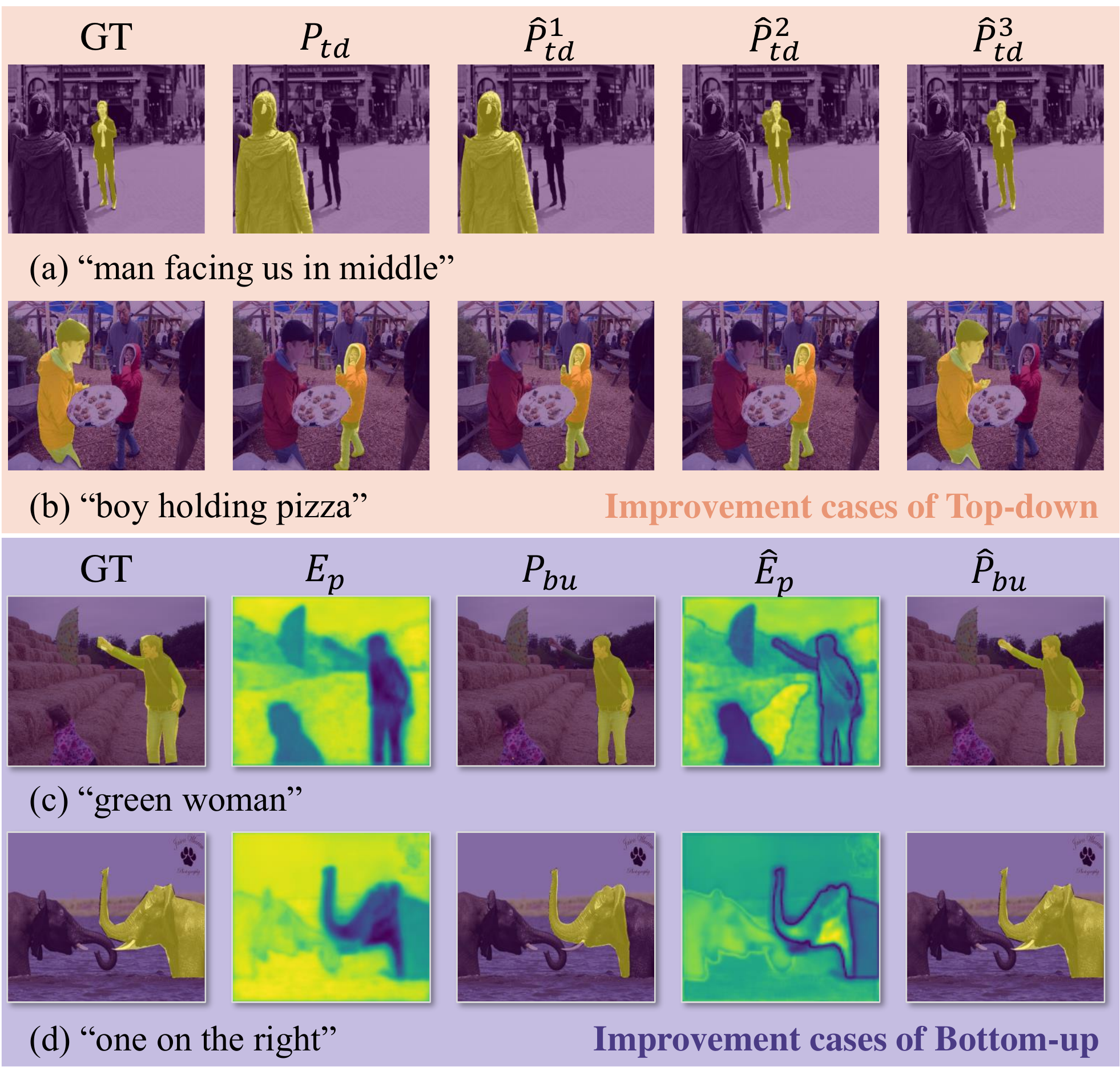}
\caption{\textbf{Qualitative effect of Complementary Feature Interaction.} 
$P_{td}$ are the original results of Top-down branch and $\hat{P}^{i}_{td}$ denotes the processed results after $i$ interaction layer. 
$E_p$ and $P_{bu}$ are the features and results of Bottom-up branch. $\hat{E}_p$ 
and $\hat{P}_{bu}$ are the features and results after Complementary Feature Interaction.}
\vspace{-5pt}
\label{fig:sup_cfi_qual}
\end{figure}

\section{Extra Qualitative Results}
\label{sup_sec:extra_qual_res}

In this section, we mainly provide extra qualitative results to show how the Complementary Feature Interaction~(CFI) improves the final results of Top-down and Bottom-up branches.

\noindent\textbf{Improvement of CFI for Top-down methods.}
For Top-down branch, CFI is described as being able to improve its robustness by introducing fine-grained information of Bottom-up branch.
Note that the feature interaction part of CFI is transformer decoder~\cite{vaswani2017attention} so that there are multiple interaction layers in CFI.
In Figure~{\color{red}{\ref{fig:sup_cfi_qual}}} (a) and (b), we compare the original results $P_{td}$ of Top-down branch to the processed results $\hat{P}_{td}^{i}$ after $i$ interaction layer of CFI.
The original failed results $P_{td}$ are fixed to correct results after several interaction layer $\hat{P}_{td}^{i}$, which shows that CFI makes the top-down results more robust by introducing fine-grained information of Bottom-up branch.

\noindent\textbf{Improvement of CFI for Bottom-up methods.}
CFI is described as being able to introduce object-centric information of Top-down branch for enhancing Bottom-up features and results.
In Figure~{\color{red}{\ref{fig:sup_cfi_qual}}} (c) and (d), we illustrate the original features $E_p$ and results $P_{bu}$ of Bottom-up branch and also visualize the processed features $\hat{E}_p$ and results $\hat{P}_{bu}$ after CFI.
The processed features $\hat{E}_p$ have clearer object edges than original features $E_p$ and the processed results $\hat{P}_{bu}$ are more precise than original results $P_{bu}$, which is concluded that CFI indeed enhances features and results of Bottom-up branch by introducing the object-centric information of Top-down branch.

\end{document}